\documentclass[]{innovator}
\usepackage{enumitem}
\usepackage[toc,page,header]{appendix}
\usepackage[utf8]{inputenc}
\usepackage[T1]{fontenc}
\usepackage{hyperref}
\usepackage{url}
\usepackage{xurl} 
\usepackage{graphicx}
\usepackage{booktabs}
\usepackage{amsfonts}
\usepackage{nicefrac}
\usepackage{makecell}
\usepackage{microtype}
\usepackage{amsmath}
\usepackage{etoolbox}
\usepackage{lipsum}  
\usepackage{minitoc}
\usepackage{tablefootnote}
\usepackage{threeparttable}
\usepackage{wrapfig}
\usepackage{appendix}
\usepackage{multirow}
\usepackage{ulem}
\useunder{\uline}{\ul}{}
\usepackage{colortbl}
\usepackage{longtable}
\usepackage{color}
\usepackage{CJKutf8}

\usepackage{adjustbox}

\title{Innovator-VL: A Multimodal Large Language Model for Scientific Discovery} 

\let\originaltitlelist\titlelist
\renewcommand{\titlelist}{%
  \centering
  \raisebox{-0.9em}{\includegraphics[width=0.06\linewidth]{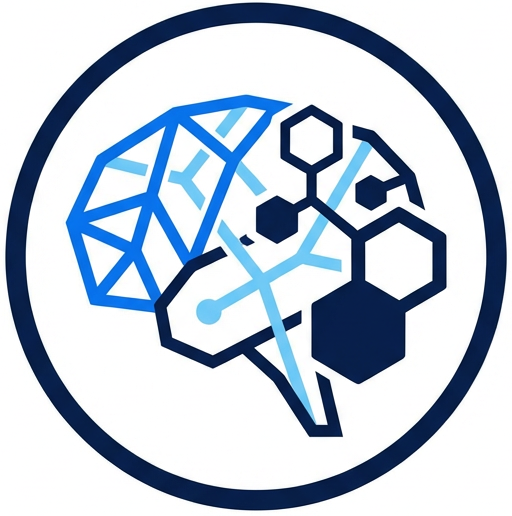}}\hspace{0.5em}%
  \originaltitlelist%
}

\def\huggingface{\raisebox{-1.5pt}{\includegraphics[height=1.05em]{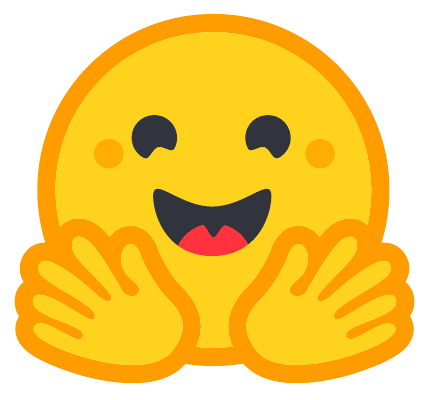}}}
\def\github{\raisebox{-1.5pt}{\includegraphics[height=1.1em]{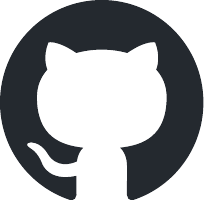}}}

\def\homepage{\raisebox{-1.5pt}{\includegraphics[height=1.15em]{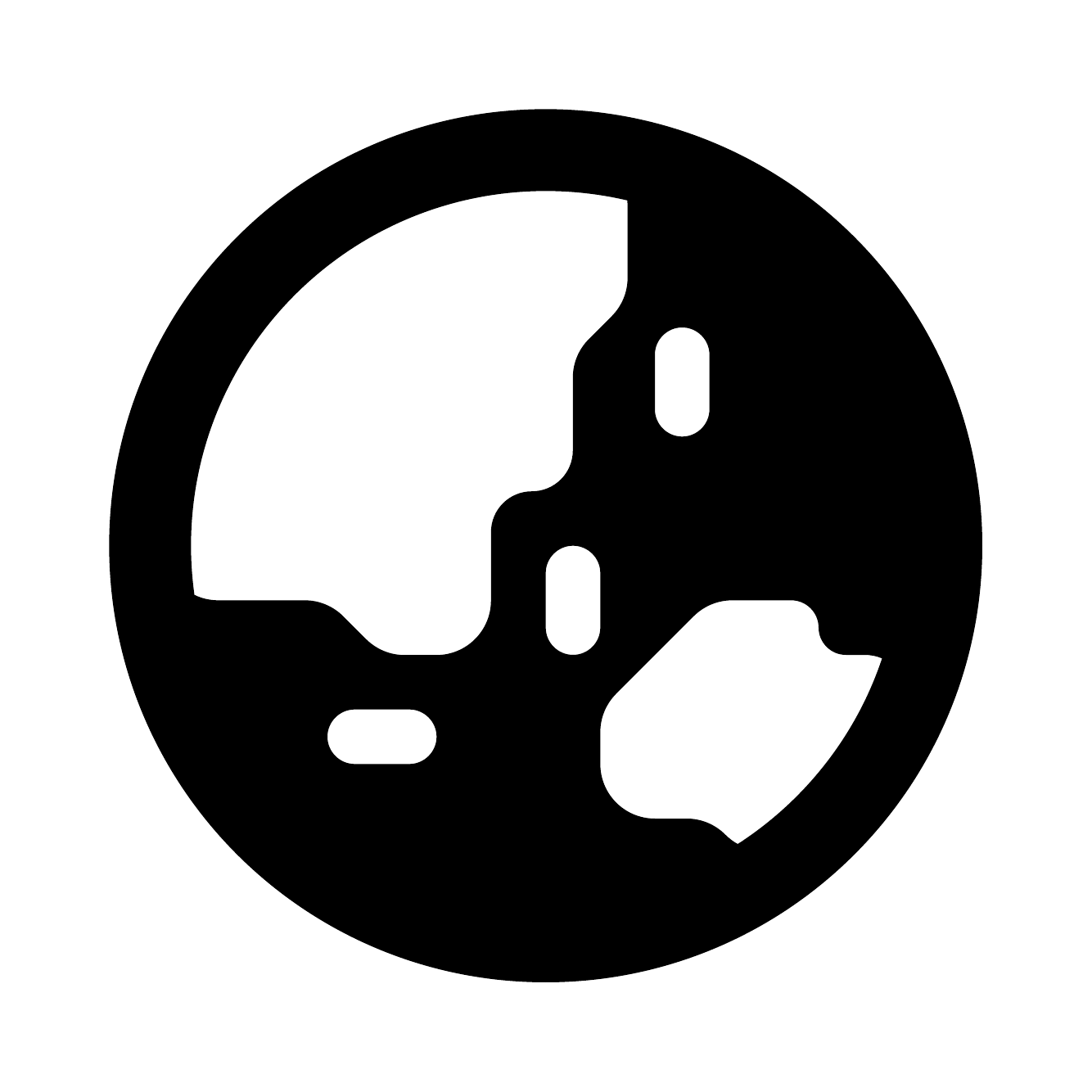}}}

\newcommand{\authorfnmark}{}
\protected\def\authorfn#1{\gappto\authorfnmark{\protect\footnotemark[#1]}}
\DeclareRobustCommand{\authorformat}[2][]{%
  \authorfont {\normalfont\bfseries \gdef\authorfnmark{}#2$^{#1}$\authorfnmark}%
}

\author[1]{Zichen Wen\authorfn{1}}
\author[1]{Boxue Yang\authorfn{1}}
\author[1]{Shuang Chen}
\author[1]{Yaojie Zhang}
\author[1]{Yuhang Han}
\author[1]{Junlong Ke}
\author[1]{Cong Wang}
\author[1]{Yicheng Fu}
\author[1]{Jiawang Zhao}
\author[1]{Jiangchao Yao}
\author[2]{Xi Fang}
\author[2]{Zhen Wang}
\author[2]{Henxing Cai}
\author[2]{Lin Yao}
\author[2]{Zhifeng Gao}
\author[2]{Yanhui Hong}
\author[2]{Nang Yuan}
\author[2]{Yixuan Li}
\author[2]{Guojiang Zhao}
\author[2]{Haoyi Tao}
\author[2]{Nan Wang}
\author[2]{Han Lyu}
\author[2]{Guolin Ke}
\author[3]{Ning Liao}
\author[3]{Xiaoxing Wang}
\author[3]{Kai Chen}
\author[3]{Zhiyu Li}
\author[3]{Feiyu Xiong}
\author[4]{Sihan Hu}
\author[4]{Kun Chen}
\author[1]{Yanfeng Wang}
\author[1]{Weinan E\authorfn{2}}
\author[2]{Linfeng Zhang\authorfn{2}}
\author[1]{Linfeng Zhang\authorfn{2}}

\affiliation[1]{School of Artificial Intelligence, Shanghai Jiao Tong University}
\affiliation[2]{DP Technology}
\affiliation[3]{MemTensor}
\affiliation[4]{Institute of Theoretical Physics, Chinese Academy of Sciences}

\abstract{
    We present Innovator-VL, a scientific multimodal large language model (MLLM) designed to advance multimodal understanding and reasoning across diverse scientific domains while still maintaining excellent performance on general vision tasks.
    Contrary to the prevailing trend of relying on massive domain-specific pretraining data and opaque training pipelines, our work demonstrates that principled training design and transparent methodology can yield strong scientific multimodal intelligence with substantially reduced data requirements.
    (i) First, we provide a fully transparent and end-to-end reproducible training pipeline for scientific multimodal modeling, covering all stages from data collection and cleaning to preprocessing, supervised fine-tuning, reinforcement learning, and evaluation, together with detailed optimization and hyperparameter recipes. This enables faithful reproduction of our results and facilitates systematic extension and adaptation by the community.
    (ii) Second, Innovator-VL exhibits remarkable data efficiency, achieving competitive performance on a wide range of scientific tasks using fewer than five million carefully curated scientific training samples, despite not relying on large-scale scientific pretraining. These results highlight that effective scientific multimodal reasoning can be achieved through principled data selection and training strategies rather than indiscriminate data scaling.
    (iii) Third, Innovator-VL demonstrates strong generalization beyond scientific domains, achieving competitive performance among MLLMs of comparable size on general vision benchmarks, multimodal reasoning benchmarks, and scientific benchmarks. This indicates that scientific alignment can be integrated into a unified multimodal model without compromising general-purpose capabilities.
    Together, our practices suggest that, in the absence of large-scale scientific data, efficient, reproducible, and high-performing scientific multimodal models can be built, thereby providing a practical and transparent foundation for future research in scientific multimodal modeling.

    \vspace{1em}
    \small \homepage~\textbf{Homepage}: \href{https://innovatorlm.github.io/Innovator-VL}{InnovatorLM.github.io/Innovator-VL} \\
    \small \github~\textbf{Github}: \href{https://github.com/InnovatorLM/Innovator-VL}{InnovatorLM/Innovator-VL} \\
    \small \huggingface~\textbf{Instruct Model}: \href{https://huggingface.co/InnovatorLab/Innovator-VL-8B-Instruct}{InnovatorLab/Innovator-VL-8B-Instruct} \\
    \small \huggingface~\textbf{Thinking Model}: \href{https://huggingface.co/InnovatorLab/Innovator-VL-8B-Thinking}{InnovatorLab/Innovator-VL-8B-Thinking} \\
    \small \huggingface~\textbf{Instruct Data}: \href{https://huggingface.co/datasets/InnovatorLab/Innovator-VL-Instruct-46M}{InnovatorLab/Innovator-VL-Instruct-46M} \\
    \small \huggingface~\textbf{RL Data}: \href{https://huggingface.co/datasets/InnovatorLab/Innovator-VL-RL-172K}{InnovatorLab/Innovator-VL-RL-172K}
    \vspace{-3em}
}

\usepackage{xcolor}
\usepackage[most]{tcolorbox}

\definecolor{ModelBox}{HTML}{F2F2F2}
\definecolor{QuestionBox}{HTML}{F2F2F2}
\definecolor{PromptBlue}{HTML}{4F8DF7}
\definecolor{LightGray}{HTML}{F2F2F2}
\definecolor{FrameGray}{HTML}{CFCFCF}

\tcbset{
  boxrule=0.8pt,
  arc=2pt,
  left=10pt,right=10pt,top=8pt,bottom=8pt,
  before skip=10pt, after skip=10pt,
  fonttitle=\bfseries\large,
}

\newtcolorbox{QuestionBox}[1]{
  breakable,
  colframe=PromptBlue,
  colback=white,
  title=#1,
}

\newtcolorbox{ModelBox}[1]{
  colframe=PromptBlue,
  colback=white,
  title=#1,
}

\newtcolorbox{UserPromptBox}[1]{
  colframe=PromptBlue,
  colback=white,
  title=#1,
}

\newtcolorbox{ThinkingBox}[1]{
  colframe=FrameGray,
  colback=white,
  title=#1,
  breakable,
}

\newtcolorbox{AnswerBox}[1]{
  colframe=PromptBlue,
  colback=white,
  title=#1,
}

\newtcolorbox{TemplateBox}[1]{
  colframe=PromptBlue,
  colback=white,
  title=#1,
}

\newtcolorbox{MyPromptBox}[1]{
  breakable,                
  colframe=PromptBlue,      
  colback=white,            
  title=#1,                 
}

\begin{document}
\maketitle

{
\footnotetext[1]{Equal contribution.}
\footnotetext[2]{Corresponding author.}
}

\newpage
\begin{spacing}{1.2}
\begingroup
\hypersetup{linkcolor=black}
\renewcommand{\cftsecfont}{\normalfont\color{black}}
\renewcommand{\cftsecpagefont}{\normalfont\color{black}}
\renewcommand{\cfttoctitlefont}{\normalfont\bfseries\sectionfont\color{black}}
\tableofcontents
\endgroup
\end{spacing}

\newpage
\begin{figure}[!t]
    \centering
    \includegraphics[width=\linewidth]{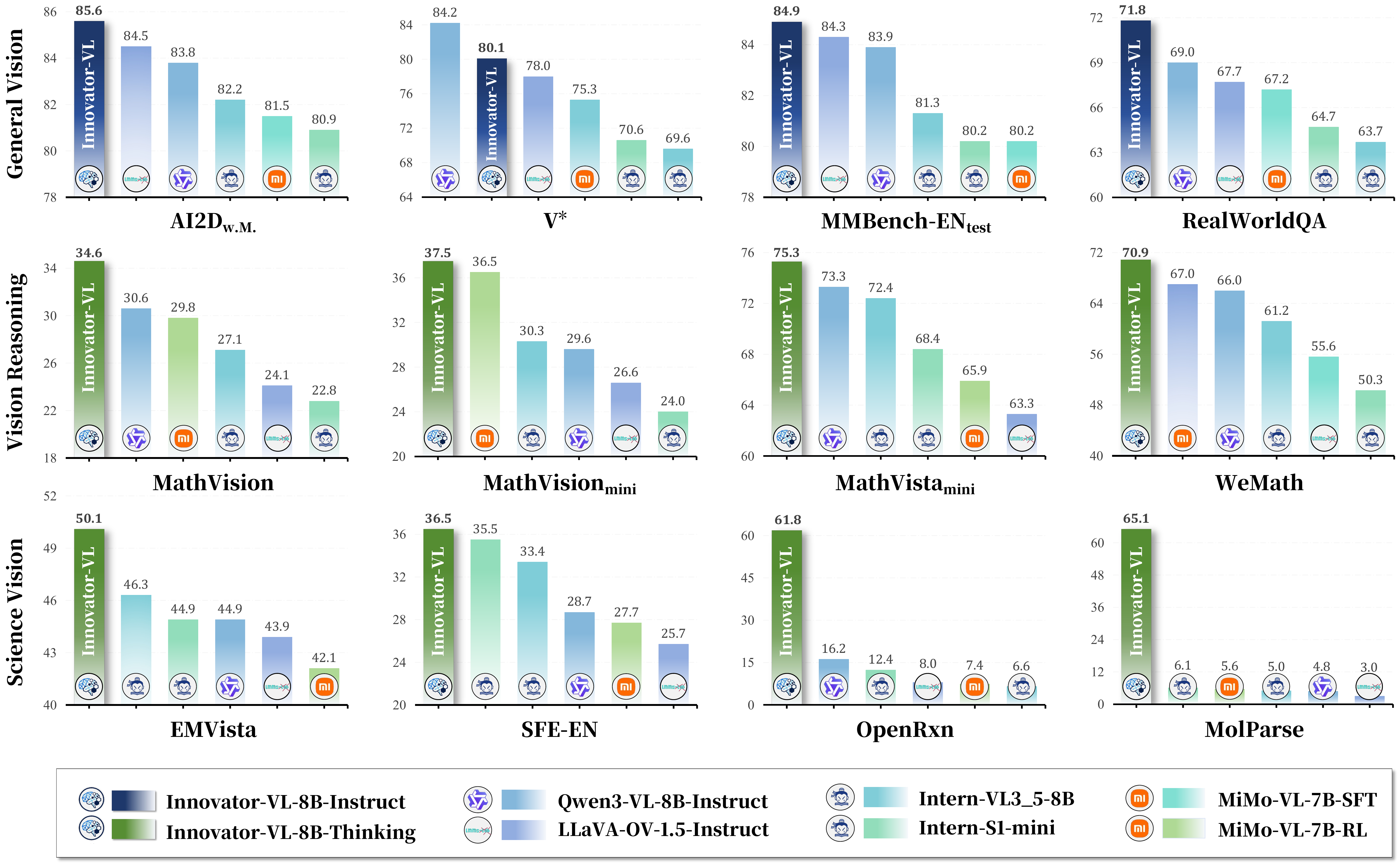}
    \caption{\textbf{Innovator-VL-8B performance across different benchmarks.} The first row displays the results of the \textbf{Innovator-VL-8B-Instruct} model on general benchmarks, while the second and third rows present the \textbf{Innovator-VL-8B-Thinking} model on mathematical reasoning and scientific benchmarks. The yellow dashed line represents the average score of the selected models.}
    \label{fig:bar_figure}
    \vspace{-0.5em}
\end{figure}
\section{Introduction}
\label{sec:introduction}
In recent years, multimodal large language models (MLLMs) have advanced rapidly~\citep{bai2025qwen2,an2025llava,wang2024qwen2,team2025kimi,team2025kimi-vl,zhu2025internvl3}, driven by the scaling of model parameters~\citep{liu2024deepseek,team2025kimi_k2}, the availability of large-scale multimodal data, and improvements in training paradigms. By jointly modeling visual and textual information, these models are able to perceive and reason, enabling a wide range of applications that were previously difficult to address with unimodal systems. As a result, MLLMs have demonstrated strong performance on diverse tasks, including visual question answering, image captioning, visual grounding, document understanding, and multimodal reasoning, highlighting their growing importance as general-purpose intelligent systems~\citep{wen2025ai}.

Despite these impressive advances, scientific-domain tasks remain particularly challenging and underexplored~\citep{zhang2023artificial}. Scientific problems often require precise understanding, rigorous multi-step reasoning, and the ability to integrate domain knowledge across modalities, making them substantially more demanding than general multimodal tasks. Importantly, strong performance on scientific tasks is widely regarded as a crucial milestone toward building artificial general intelligence (AGI) and scientific general intelligence (SGI), as it reflects a model's capacity for systematic reasoning and knowledge-intensive problem solving. However, existing multimodal large language models, especially open-source models, still struggle to achieve performance in scientific tasks comparable to that observed on general-purpose benchmarks, revealing a significant gap between current capabilities and the requirements of scientific reasoning.

Recognizing this challenge, researchers in both academia and industry have begun to devote increasing attention to scientific multimodal understanding and reasoning, leading to a number of recent advances. Several scientific-oriented multimodal large language models~\citep{bai2025intern,zhong2024fuxi,liao2025innovator} and benchmarks~\citep{li2025rxnbench,zhao2025superchem,shen2024fine} have been proposed to enhance performance on tasks involving mathematics, physics, chemistry, and other STEM domains. These efforts demonstrate the potential of multimodal models in scientific problem solving and represent important steps toward bridging the gap between general-purpose perception and scientific domain-specific reasoning. While these efforts have led to encouraging progress, many existing approaches rely on large-scale or highly specialized scientific datasets, making them expensive to reproduce and difficult to extend. In addition, the training pipelines of several systems are either partially opaque or tightly coupled with task-specific heuristics, limiting reproducibility and broader adoption by the community. Moreover, improvements on scientific benchmarks often come at the cost of degraded performance on general multimodal understanding and reasoning tasks, suggesting a lack of balanced generalization.

Motivated by these observations, we present \textbf{Innovator-VL}, a scientific multimodal large language model designed to advance multimodal understanding and reasoning across diverse scientific domains. In contrast to prior approaches that depend on massive domain-specific data or opaque training pipelines, Innovator-VL adopts a principled and fully transparent training framework that emphasizes data efficiency, reproducibility, and balanced generalization. 
Specifically, we use RICE-ViT~\citep{xie2025region} as the visual encoder for Innovator-VL.
Scientific understanding in multimodal settings critically depends on the accurate perception of structured visual elements, such as symbols, annotations, and relational components. By capturing fine-grained, region-level semantics, RICE-ViT decomposes scientific images into semantically coherent visual units, enabling faithful alignment between visual entities and scientific concepts. This structured visual representation significantly reduces the burden on downstream language reasoning modules, facilitating more reliable scientific reasoning.
Following the design of the QwenVL series~\citep{wang2024qwen2,bai2025qwen2}, we adopt PatchMerger as the vision-language projector. This design effectively balances representational capacity and computational efficiency by merging visual patches into a compact yet semantically informative representation. 
Considering that Qwen3-8B-Base~\citep{yang2025qwen3} has been extensively pre-trained on a broad and diverse corpus, we adopt it as the language decoder for Innovator-VL.

During \textbf{Pre-training}, we focus on robust language-image alignment and high-quality knowledge injection using LLaVA-1.5 558k~\citep{liu2024improved} and LLaVA-OneVision-1.5-Mid-Training-85M~\citep{an2025llava}. 
Given that Qwen3-8B-Base has already undergone extensive pretraining and high-quality scientific-domain data is unavailable, we did not further pretrain the language model on scientific texts or include scientific data in the multimodal pretraining stage. 
In the \textbf{Supervised Fine-tuning (SFT)} stage, we prioritize data quality over quantity. We construct high-quality, human-in-the-loop synthetic data pipelines for multiple specialized scientific domains (e.g., in-the-wild OCSR, chemical reaction understanding, and microstructural characterization). Using less than 5 million scientific samples, Innovator-VL achieves competitive performance among models of comparable size. Meanwhile, we incorporate data with chain-of-thought and multi-step reasoning to prepare the model for solving complex scientific problems and the subsequent RL stage.
Finally, to bridge the gap between the model's latent reasoning capabilities and consistent performance, we introduce \textbf{Reinforcement Learning (RL)}. We curate a discrepancy-driven RL dataset and employ Group Sequence Policy Optimization (GSPO)~\citep{zheng2025group} to optimize long-horizon reasoning paths, guided by a hierarchical reward system that enforces both structural format and semantic accuracy.

In summary, our work and practice make three main contributions:
\begin{itemize}
    \item We present \textbf{Innovator-VL}, a fully transparent and end-to-end reproducible training recipe for scientific multimodal modeling, facilitating further exploration of scientific multimodal large language models by the research community.

    \item Innovator-VL demonstrates \textbf{remarkable data efficiency}, achieving competitive performance across a wide range of scientific tasks using fewer than \textbf{five million} carefully curated high-quality scientific samples, despite not relying on large-scale scientific-domain pretraining.
    
    \item Innovator-VL exhibits \textbf{strong capability beyond scientific domains}, attaining competitive performance among multimodal large language models of comparable size on general vision benchmarks, multimodal reasoning benchmarks, and scientific benchmarks. This indicates that scientific reasoning enhancement can be effectively incorporated into multimodal models without compromising general-purpose capabilities.

  \end{itemize}

\section{Model Architecture}
\label{sec:arch}

\begin{figure}[!t]
    \centering
    \includegraphics[width=\linewidth]{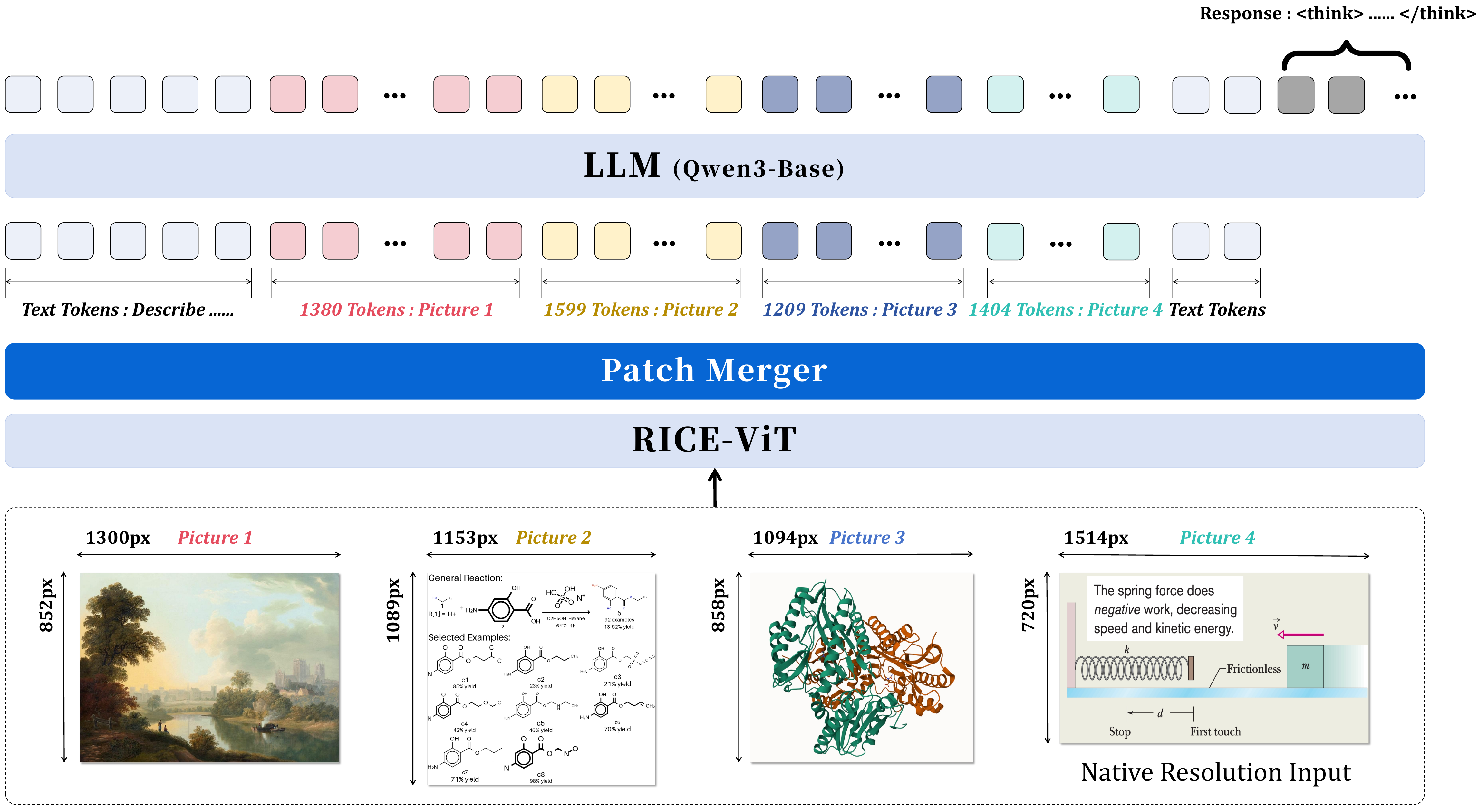}
    \caption{\textbf{Overall architecture of our native-resolution multi-image reasoning model.} Given a text prompt and multiple images with heterogeneous resolutions, \textbf{RICE-ViT} encodes each image at its native size, producing variable-length visual token sequences. \textbf{Patch Merger} then aggregates the visual tokens into a compact sequence that is concatenated with text tokens and fed into \textbf{Qwen-8B-base} to generate the final response with explicit reasoning.}
    \label{fig:architecture}
\end{figure}

Innovator-VL adopts a three-stage architecture following the widely adopted paradigm of \textit{vision encoder-projector-language model}, which has been shown to effectively bridge visual and textual modalities. Our model architecture is illustrated in Figure~\ref{fig:architecture}.

\subsection{Vision Encoder}
We adopt \textbf{RICE-ViT}~\citep{xie2025region}, a Vision Transformer variant designed for region-aware representation learning. Unlike CLIP~\citep{radford2021learning} or SigLIP~\citep{zhai2023sigmoid,tschannen2025siglip} that primarily focus on global patch interactions, RICE-ViT integrates region-level semantics through a specialized Region Transformer layer, enabling the model to capture both holistic and localized visual cues within a single forward pass. This is particularly important for scientific imagery, where fine-grained structures, dense labels, and spatially localized patterns frequently occur. The region-aware cluster discrimination mechanism used in RICE-ViT facilitates richer semantic embedding by jointly optimizing object and OCR region representations, which improves downstream tasks such as grounding, dense prediction, and multimodal reasoning. Existing studies have shown that RICE-ViT consistently outperforms traditional visual backbones across diverse visual tasks, including segmentation and dense detection. By integrating RICE-ViT as our vision encoder, Innovator-VL benefits from enhanced visual representation capacity that is well aligned with the demands of scientific multimodal understanding and reasoning.

\subsection{Projector}
To bridge the visual encoder and language model, Innovator-VL employs a learned token compression module, namely \textbf{PatchMerger}, following the design used in recent high-performance multimodal language models~\citep{wang2024qwen2,bai2025qwen2}. The PatchMerger module is specifically introduced to address the quadratic computational cost associated with processing dense visual tokens from the Vision Transformer. By learning to merge a larger set of input patch embeddings into a smaller set of representative tokens, PatchMerger significantly reduces the token sequence length that downstream components must process, leading to substantial reductions in computational complexity and memory usage. Empirical studies on token compression~\citep{liao2025we,wen2025token} have demonstrated that such learned merger modules can achieve significant efficiency gains with negligible accuracy degradation on both upstream and downstream tasks~\citep{zhang2025d2pruner,liu2025shifting,han2024filter,wen2025stop,wen2025efficient,liu2025global}. In the context of multimodal integration, this compression facilitates more efficient cross-modal interaction with the language model without compromising the richness of visual representations, which is particularly important when training on limited scientific datasets. By adopting PatchMerger as the projector, Innovator-VL achieves a favorable balance between computational efficiency and representational fidelity, enabling scalable multimodal reasoning and fine-grained understanding.

\subsection{Language Model}
For the language modeling and reasoning component of Innovator-VL, we adopt \textbf{Qwen3-8B-Base}~\citep{yang2025qwen3}. Pretrained on a large and diverse corpus spanning multiple domains and languages, Qwen3-8B-Base exhibits strong performance in STEM, logical reasoning, and long-context understanding, making it particularly suitable for scientific multimodal tasks that require integrating textual and visual information. Furthermore, Qwen3-8B-Base is open-source and supported by a mature tooling ecosystem, which aligns with our goal of a fully transparent and reproducible scientific multimodal large language model. By integrating Qwen3-8B-Base as the language backbone, Innovator-VL is able to perform advanced scientific reasoning, multimodal question answering, and descriptive generation.

\begin{figure}[!t]
    \centering
    \includegraphics[width=\linewidth]{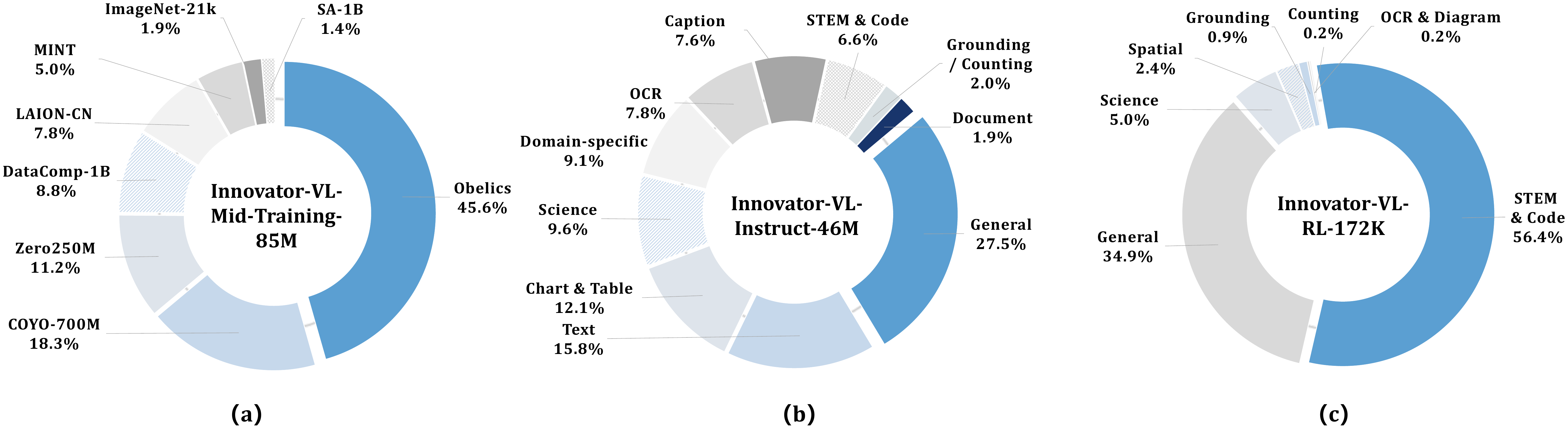}
    \caption{\textbf{Data distribution across different training stages.} (a) Distribution of data sources within the Innovator-VL-Mid-Training-85M dataset. (b) Distribution of data sources within the Innovator-VL-Instruct-46M dataset. (c) Distribution of data sources within the Innovator-VL-RL-172K dataset.}
    \label{fig:pie_figure}
\end{figure}

\section{Pre-training}
\label{sec:pretraining}

Our pre-training consists of two stages.
\begin{itemize}
    \item \textbf{Language-Image Alignment}. We pretrain the projector to align visual features with the word embedding space of large language models (LLMs).
    \item \textbf{High-quality Mid-Training}. We introduce a high-quality knowledge learning (Mid-training) stage to balance computational efficiency with the effective injection of new knowledge into multimodal large language models (MLLMs). In this stage, we transition to full-parameter training across all modules.
\end{itemize}

\begin{table*}[htbp]
    \centering
    \caption{Overview of training stages, trainable components, datasets, and data scales used to train Innovator-VL, where Innovator-VL-Mid-Training-85M is derived from LLaVA-OneVision-1.5-Mid-Training-85M.}
    \label{tab:training_setup}
    \begin{tabular}{@{}llcc@{}}
    \toprule
    \textbf{Stage} & \textbf{Training} & \textbf{Dataset} & \textbf{Samples} \\ \midrule
    Language-Image Alignment & Projector & LLaVA-1.5 558k & \textasciitilde 558K \\
    High-quality Mid-Training & All & Innovator-VL-Mid-Training-85M & \textasciitilde 85M \\
    Supervised Fine-tuning & All & Innovator-VL-Instruct-46M & \textasciitilde 46M \\
    Reinforcement Learning & All & Innovator-VL-RL-172K & \textasciitilde 172K \\ \bottomrule
    \end{tabular}
\end{table*}

Given the scarcity and annotation difficulty of large-scale, high-quality scientific training data, and considering that Qwen3-8B-Base has already undergone extensive pretraining on a comprehensive and diverse corpus, we chose not to continue pretraining the LLM on scientific text data\footnote{We have also developed an enhanced internal variant based on a Qwen3-8B-Base that underwent continued pretraining on high-quality scientific corpora. Due to ongoing data compliance review procedures, this version is not yet available for public release.}. We reason that Qwen3-8B-Base has already absorbed substantial scientific knowledge during its pretraining. Without access to high-quality and diverse scientific data, further pretraining on scientific texts could introduce bias and overfitting to some domains. At the same time, our goal is for Innovator-VL to excel in scientific tasks while maintaining its general capabilities, with a robust and reliable foundation in visual understanding serving as the cornerstone for higher-level scientific reasoning. 

This philosophy extends naturally to the multimodal pretraining stage. In the first stage, \textbf{Language-Image Alignment}, we train our projector using the LLaVA-1.5 558k~\citep{liu2024improved} dataset, aiming to align visual features with the word embedding space of the LLM. 

Building upon this alignment stage, we introduce the \textbf{High-quality Mid-Training}. In this stage, we transition to full-parameter training across all modules. 
We employ the \textbf{LLaVA-OneVision-1.5-Mid-Training} dataset~\citep{an2025llava}, a large-scale multimodal corpus specifically designed to enhance the foundation ability of multimodal large language models (MLLMs). This dataset comprises approximately \textbf{85 million} high-quality image-text pairs, including 65 million English and 20 million Chinese samples. The data are curated from a diverse array of prominent sources, including COYO-700M~\citep{coyo700m}, Obelics~\citep{laurenccon2023obelics}, DataComp-1B~\citep{gadre2023datacomp}, LAION-CN~\citep{zhang2022fengshenbang}, ImageNet-21K~\citep{russakovsky2015imagenet}, SAM-1B~\citep{kirillov2023segment}, MINT~\citep{wang2023mint}, and Zero250M~\citep{xie2023ccmb}.
A key characteristic of this dataset is the integration of a \textbf{feature-based concept-balanced sampling strategy}. To address the challenges of missing annotations or suboptimal caption quality (e.g., the brief and incomplete labels in COYO-700M or caption-free nature of SAM-1B), the dataset construction process moves beyond traditional raw text matching. Instead, it utilizes the pre-trained \textbf{MetaCLIP-H/14-Full-CC2.5B}~\citep{xu2023demystifying} encoders to project both images and a taxonomy of 500,000 concepts into a shared embedding space. By retrieving the top-$K$ nearest concept embeddings for each image, refined \textbf{pseudo-captions} are generated. This methodology ensures semantic diversity and balance across the massive-scale data, ultimately facilitating a more robust and fine-grained visual-language alignment during pre-training.
\section{Post-training}
\label{sec:posttraining}
Following multimodal pretraining, Innovator-VL undergoes staged post-training in order to further refine the model's capabilities, including supervised fine-tuning and reinforcement learning.
This stage aims to enhance the model's ability to follow instructions, perform multi-step reasoning for complex problems, and conduct multimodal reasoning in scientific scenarios.

\subsection{Supervised Fine-tuning}
To further empower the model to address a wide spectrum of visual tasks with precise and appropriate responses, we continue to perform full-parameter visual instruction fine-tuning on the model.
At this stage, we pursue three objectives. First, we endow the model with fundamental visual instruction-following capability, enabling it to leverage the knowledge acquired during pretraining to solve general multimodal understanding tasks. Second, to prepare for the subsequent reinforcement learning stage, we perform a cold start by equipping the model with preliminary chain-of-thought and multi-step reasoning abilities. Third, and most importantly, we strengthen the model's scientific understanding, training it to effectively solve problems in scientific domains. To this end, we curate three categories of training data for supervised fine-tuning (SFT), including general multimodal instruction data, chain-of-thought and multi-step reasoning data, and scientific understanding data.

\subsubsection{General Multimodal Instruction Data}
We employ the LLaVA-OneVision-1.5-Instruct dataset~\citep{an2025llava}, which consolidates a diverse collection of instruction-tuning data from multiple sources. The dataset is carefully curated to ensure balanced coverage across a wide range of categories, including Caption, Chart \& Table, Code \& Math, General VQA, Grounding \& Counting, OCR. This dataset contains approximately 22 million samples. The strong performance of LLaVA-OneVision-1.5 on general vision benchmarks empirically demonstrates the effectiveness of this data.

\subsubsection{Chain-of-Thought and Multi-step Reasoning Data}
Compared to general multimodal tasks, scientific tasks are inherently more challenging. Therefore, in the visual instruction fine-tuning stage, we not only employ general multimodal instruction data but also introduce data with chain-of-thought and multi-step reasoning annotations. This design enables the model to acquire reasoning patterns at this stage, laying a solid foundation for tackling complex scientific reasoning tasks as well as for the subsequent reinforcement learning stage.
Concretely, we adopt a large-scale reasoning-oriented instruction dataset Honey-Data-15M~\citep{zhang2025beehighqualitycorpusfullstack} containing approximately 15 million samples. The dataset spans a wide range of domains and is specifically designed to elicit chain-of-thought and multi-step reasoning behaviors. Different from the original dataset, we further refine the data by removing a large number of explicit \texttt{think} tags. We find that such heavily templated markers may introduce undesirable noise during training; their removal helps preserve the underlying reasoning structure while encouraging more natural and robust reasoning patterns.

\subsubsection{Scientific Understanding Data}

\begin{figure}[!t]
    \centering
    \includegraphics[width=\linewidth]{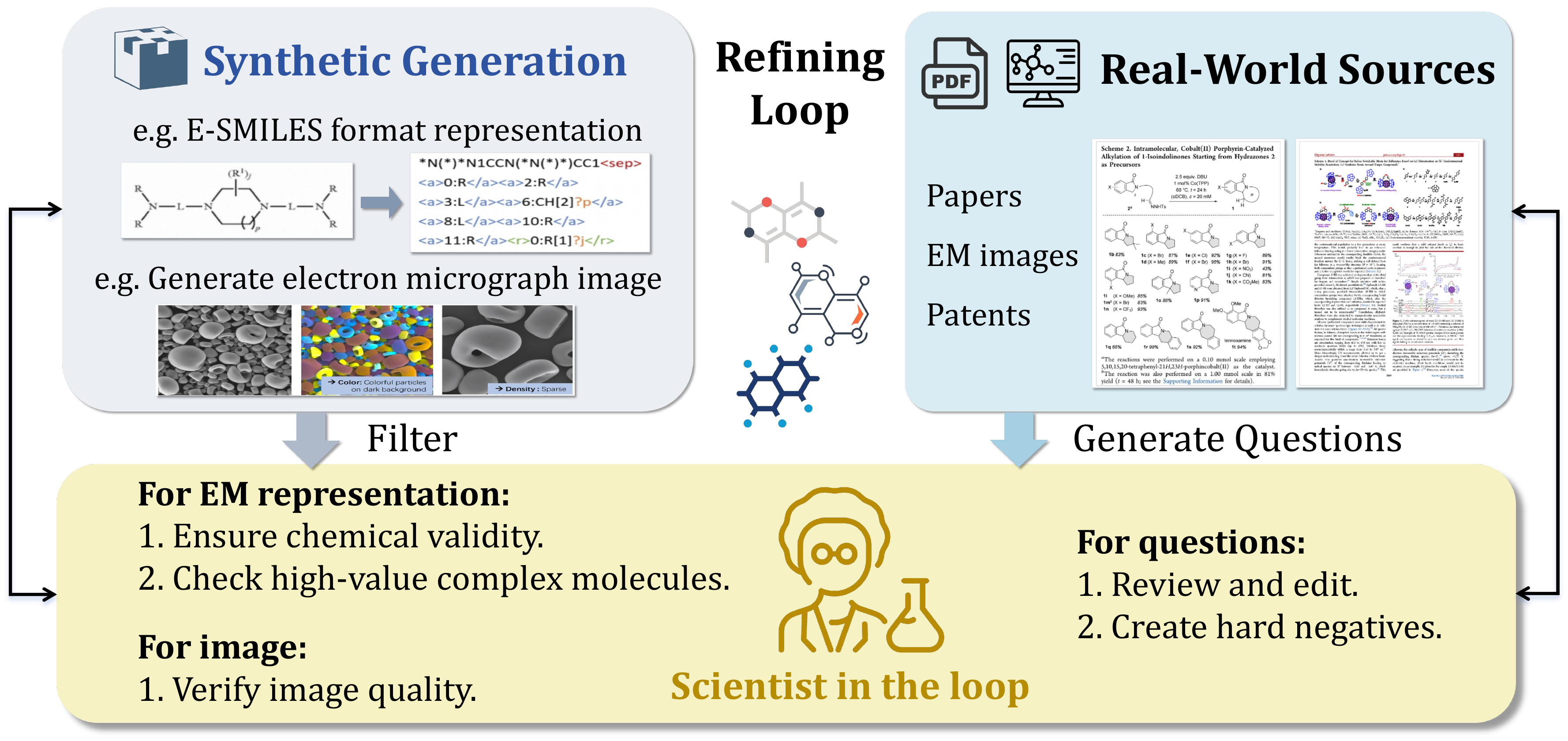}
    \caption{\textbf{Data Construction Pipeline.} The raw data are obtained from both synthetic generation and real-world sources. For each data modality (EM representations, images, and questions), domain experts apply modality-specific inspection and refinement strategies. Through repeated iterative optimization, a final high-quality dataset is produced.}
    \label{fig:data_pipeline}
\end{figure}
To enhance the model's generalizable understanding and cross-domain robustness on scientific tasks, we synthesize high-quality datasets across diverse scientific domains. Our data construction philosophy emphasizes principled methodology, systematic quality control, and domain-specific refinement strategies. To illustrate this approach, we present three representative scientific subfields as exemplars: in-the-wild optical chemical structure recognition (OCSR), chemical reaction understanding from scientific literature, and microstructural characterization from electron micrographs (EM). For each subfield, we develop a dedicated data construction and quality control pipeline that systematically integrates data generation, filtering, and consistency verification, demonstrating our generalizable framework for scientific data synthesis. The overall data processing pipeline is illustrated in Figure~\ref{fig:data_pipeline}.

\smallskip
\noindent\textbf{Optical chemical structure recognition (in-the-wild OCSR).}
A low-cost, large-scale paired training corpus is developed via a human-in-the-loop data engine with two stages: (i) synthetic bootstrapping; (ii) active-learning-driven expansion on real-world data. To represent patent-prevalent yet SMILES-inexpressible structures (e.g., Markush patterns and connectivity constraints), E-SMILES is adopted as a unified annotation format, written as \texttt{SMILES$\langle$sep$\rangle$EXTENSION}, where \texttt{EXTENSION} uses compact XML-like tokens to encode R-groups and connectivity information while remaining RDKit-compatible and sequence-model-friendly. The initial model is trained on $\sim$7M synthesized and rendered paired molecular structure images--E-SMILES samples, and subsequently expanded with real patent/paper PDFs through detection, cropping, and deduplication. Confidence is estimated via a 5-fold ensemble and SMILES-similarity scoring, and mid-confidence samples (0.6--0.9) are prioritized for expert correction. The pipeline iterates in a closed loop of ``model pre-annotation--human verification--periodic retraining'', enabling scalable acquisition of high-quality real annotations and progressive alignment to real-world distributions.

\smallskip
\noindent\textbf{Chemical reaction understanding from scientific literature.}
To systematically evaluate multimodal large language models (MLLMs) in realistic PDF-based reaction-reading scenarios, a hierarchical benchmark is curated to cover both single-figure fine-grained perception and full-document cross-modal reasoning. High-impact open-access papers from the past five years are selected to ensure broad reaction diversity and authentic document complexity. Reaction schemes are extracted via PDF layout parsing and automatic cropping (retaining titles and label regions). QA pairs are generated in a human-in-the-loop manner: candidates are proposed by a large model and subsequently reviewed and refined by domain experts to ensure chemical consistency and verifiability. Expert-driven adversarial distractor design further yields chemically plausible, visually dependent hard negatives, while verification mechanisms such as ``None of the Above'' are incorporated to suppress guess-based hallucinations and improve benchmark robustness.

\smallskip
\noindent\textbf{Microstructural characterization from electron micrographs (EM).}
The EM dataset is assembled at scale with dense instance-level segmentation annotations and structured textual descriptions to support microstructure segmentation and generation. Multi-source real EM data are aggregated and cleaned via deduplication and quality filtering, while non-structural regions (e.g., watermarks and instrument text) are cropped to obtain a cleaner and more consistent visual distribution. A nine-dimensional, attribute-disentangled description schema is introduced, leveraging multi-model independent extraction with cross-validation followed by expert adjudication for accuracy and interpretability. Instance segmentation annotations are produced through a high-performance annotation platform and a human-in-the-loop iterative workflow, forming a closed loop of ``expert seed labeling--model pseudo-label generation--human refinement--multi-stage quality control'' to enable scalable construction of high-confidence annotations.

\subsection{Reinforcement Learning}
To further strengthen the model's multimodal reasoning and its capacity to tackle complex scientific problems, we apply reinforcement learning (RL) to Innovator-VL-8B-Instruct.

\subsubsection{RL Data}
We curate a specialized RL training dataset, \textbf{Innovator-VL-RL-172K}. Our construction pipeline integrates discrepancy-driven selection, reward-based filtering, and rigorous format standardization.
\paragraph{Discrepancy-Driven Selection and Filtering.}
We view RL primarily as an \textit{elicitation} mechanism rather than knowledge injection. Consequently, our data selection targets the model's "effective learnable boundary." We measure the divergence between \textit{Pass@N} and \textit{Pass@1} performance across candidate samples; a significant gap indicates that while the model possesses the potential to generate correct solutions (high \textit{Pass@N}), its policy fails to reliably prioritize them (low \textit{Pass@1}).
To ensure high training efficiency, we further employ a \textbf{reward-based sampling} strategy. We generate multiple response candidates for each instance and filter based on average reward scores, retaining only medium-difficulty instances. This effectively removes trivial samples the model has already mastered and unsolvable cases beyond its current grasp.
\paragraph{Standardization and Composition.}
Raw data sourced from diverse public datasources~\citep{zhang2025openmmreasoner,xu2025llava,deng2025openvlthinker,qiao2025we,an2025llava,wang2025vl,li2025unisvg} often suffers from inconsistent answer styles and reasoning structures. To mitigate training instability, we standardize all collected samples into a unified reasoning format, normalizing textual structures and ensuring consistent step-wise outputs.
The resulting curated corpus contains approximately 172K instances. As illustrated in Figure~\ref{fig:pie_figure}, the dataset is heavily weighted towards complex reasoning tasks. Specifically, \textbf{STEM \& Code} represents the majority share at 56.4\%, followed by \textbf{General} multimodal tasks at 34.9\%. To ensure comprehensive coverage of fine-grained capabilities, the dataset also includes specific subsets for \textbf{Science} (5.0\%), \textbf{Spatial} reasoning (2.4\%), \textbf{Grounding} (0.9\%), \textbf{Counting} (0.2\%), and \textbf{OCR \& Diagram} understanding (0.2\%).

\subsubsection{Optimization Algorithm}

To effectively train Innovator-VL on complex reasoning tasks requiring long Chain-of-Thought (CoT) paths, we employ \textbf{Group Sequence Policy Optimization (GSPO)}~\citep{zheng2025group}. While previous group-based methods like GRPO apply token-level importance weights to sequence-level rewards, this mismatch can introduce high-variance noise and instability during training. GSPO addresses this by strictly aligning the optimization unit with the reward unit, performing importance sampling and clipping at the sequence level.

GSPO optimizes the policy $\pi_\theta$ by maximizing the expected reward of generated response groups while constraining the update using a sequence-level trust region. The objective function is defined as:

\begin{equation}
    \mathcal{J}_{\text{GSPO}}(\theta) = \mathbb{E}_{x \sim \mathcal{D}, \{y_i\}_{i=1}^G \sim \pi_{\theta_{\text{old}}}(\cdot|x)} \left[ \frac{1}{G} \sum_{i=1}^G \min \left( s_i(\theta) \widehat{A}_i, \text{clip}(s_i(\theta), 1-\epsilon, 1+\epsilon) \widehat{A}_i \right) \right],
    \label{eq:gspo_objective}
\end{equation}

where $G$ is the group size, and $\widehat{A}_i$ is the normalized advantage for the $i$-th response derived from the reward $r(x, y_i)$. A key innovation of GSPO is the definition of the importance ratio $s_i(\theta)$, which is based on the length-normalized sequence likelihood:
\begin{equation}
    s_i(\theta) = \left( \frac{\pi_\theta(y_i|x)}{\pi_{\theta_{\text{old}}}(y_i|x)} \right)^{\frac{1}{|y_i|}}.
    \label{eq:gspo_importance_ratio}
\end{equation}

By utilizing this sequence-level objective, GSPO eliminates the inconsistency of token-level updates in previous algorithms, ensuring robust convergence and superior efficiency when scaling RL to the long-context reasoning tasks present in our dataset.

\subsubsection{Reward System}

To facilitate deep reasoning capabilities, we explicitly instruct the model to separate its internal thought process from the final conclusion. All RL training instances are formatted with the following instruction:

\noindent\fbox{%
    \parbox{0.95\linewidth}{%
        \small\texttt{\textbf{Instruction:} Think and solve the following question step by step. Please put your thinking and analysis procedure within <think>...</think>. Put ONLY your final answer within <answer>...</answer>.}
    }%
}

To ensure the generated responses are both structurally compliant and factually correct, we design a comprehensive reward function $R(x, y)$ that integrates format adherence with a hierarchical accuracy verification process. The total reward is calculated as a weighted sum:

\begin{equation}
    R(x, y) = \alpha \cdot r_{\text{format}} + (1 - \alpha) \cdot r_{\text{accuracy}}.
\end{equation}

where we empirically set the weighting factor $\alpha = 0.1$ to prioritize reasoning correctness while maintaining structural discipline.

\paragraph{Format Reward ($r_{\text{format}}$).} 
To facilitate stable Chain-of-Thought (CoT) and multi-step reasoning, we enforce a strict output structure. The model receives a format reward of $1.0$ if the response follows the specific XML-like template \texttt{<think>...</think>, <answer>...</answer>} or employs standard mathematical formatting like \texttt{\textbackslash boxed\{\}}. Partial credit (e.g., $0.8$) is assigned if the response contains mathematical reasoning and a discernible answer but lacks precise formatting tags, while completely unstructured outputs receive a score of $0.0$.

\paragraph{Accuracy Reward ($r_{\text{accuracy}}$).}
Correctness verification is the core of our reward system. Given the diversity of tasks (e.g., multiple-choice, math, open-ended QA), we employ a \textbf{hierarchical verification strategy} that balances computational efficiency with semantic understanding. The process proceeds in three cascading stages:

\begin{enumerate}
    \item \textbf{Heuristic \& Rule-Based Matching:} We first extract the answer candidate using regular expressions (targeting \texttt{<answer>} tags or \texttt{\textbackslash boxed\{\}}). For Multiple-Choice Questions (MCQs), we utilize an extensive pattern-matching algorithm to handle various output styles (e.g., ``(A)'', ``The answer is A'', ``Option A''). For general text, we apply relaxed string matching to account for minor variations in punctuation or spacing.
    
    \item \textbf{Symbolic Verification:} If exact matching fails for mathematical problems, we employ a symbolic verification tool (using \texttt{math\_verify}) to parse both the ground truth and the prediction. This allows us to verify mathematical equivalence (e.g., recognizing that $\frac{1}{2}$ is equivalent to $0.5$) which simple string matching would miss.
    
    \item \textbf{LLM-as-a-Judge:} For open-ended questions or cases where deterministic methods yield a negative result, we utilize a strong external LLM (e.g., Qwen3-VL-235B-A22B-Instruct~\citep{yang2025qwen3,chen2024mj}) as a final judge. The judge is prompted with a strict evaluation protocol to output a binary score ($0$ or $1$) based on semantic alignment with the ground truth, ignoring trivial formatting differences.
\end{enumerate}

This cascaded approach ensures that clear, deterministic answers are rewarded cheaply and quickly, while complex semantic matches are accurately captured by the model-based judge.

\section{Infrastructure}
\label{sec:infrastructure}

During the pre-training and supervised fine-tuning (SFT) stages of Innovator-VL, our training pipeline is built upon \textbf{AIAK-Training-LLM}\footnote{An enterprise-grade optimization of Megatron-LM provided by Baidu Cloud.}, a robust distributed training framework derived from Megatron-LM~\citep{shoeybi2019megatron}. 
To maximize computational throughput and memory efficiency, we employ a suite of advanced optimization strategies, including the \textbf{Transformer Engine} for accelerating mixed-precision computation, distributed optimizer parallelism, and selective activation recomputation to significantly reduce memory overhead. 
This high-performance training infrastructure serves as the backbone of our pre-training and SFT stages, enabling efficient and scalable learning over massive multimodal datasets.

Following LLaVA-OneVision-1.5~\citep{an2025llava}, we adopt a data packing strategy to further improve training efficiency. 
In large-scale multimodal training, substantial variation in sequence lengths leads to significant computational redundancy caused by excessive padding. 
To address this issue and improve GPU utilization, we employ a carefully designed offline data packing strategy. 
Specifically, instead of dynamically collating samples at runtime, we pre-optimize the data layout through a parallelized, bucket-based algorithm that groups and concatenates multiple short instances into dense, contiguous sequences. 
This global preprocessing step enables more effective utilization of the context window while minimizing padding overhead. 
As a result, our approach substantially improves training throughput and overall computational efficiency.

To overcome the efficiency bottlenecks inherent in synchronous RL training, where the optimization process is frequently blocked by time-consuming inference, we utilize \textbf{AReaL}~\citep{fu2025areal}, a high-throughput asynchronous reinforcement learning framework. By fundamentally decoupling the \textit{Rollout} phase from the \textit{Learning} phase, AReaL transforms the training pipeline into a non-blocking producer-consumer system. This architecture allows dedicated inference workers to continuously produce long-horizon reasoning trajectories, filling the replay buffer without stalling the trainer nodes. Consequently, the learner can perform gradient updates in parallel with data generation, ensuring near-100\% GPU utilization and drastically reducing the wall-clock time required for convergence on complex reasoning tasks.

\begin{table*}[!t]
\centering
\renewcommand{\arraystretch}{1.2} 
\setlength{\tabcolsep}{3.5pt} 
\scriptsize 

\newcommand{\best}[1]{{\fontseries{b}\selectfont #1}}
\caption{Comparison with state-of-the-art multimodal models across general, mathematical, and scientific benchmarks. The \best{best} and \underline{second-best} results are highlighted in bold and underline, respectively.}
\resizebox{\textwidth}{!}{
\begin{tabular}{l | cc | ccccccc} 
\toprule
\multirow{2.5}{*}{\textbf{Benchmark}} & \multicolumn{2}{c|}{\textbf{Innovator-VL}} & \textbf{Qwen3-VL} & \textbf{InternVL3.5} & \textbf{Intern-S1} & \textbf{LLaVA-OV} & \multicolumn{2}{c}{\textbf{MiMo-VL}} & \textbf{MiniCPM-V} \\
 & 8B-Instruct & 8B-Thinking & 8B & 8B & mini (9B) & 1.5-8B & 7B-SFT & 7B-RL & 4.5 (8B) \\
\midrule

\rowcolor[HTML]{EBF2FF} \multicolumn{10}{l}{\textbf{General}} \\
AI2D                                   & \best{85.56} & \underline{85.07} & 83.78 & 82.19 & 80.86 & 84.49 & 81.54 & 81.51 & 83.42 \\
AI2D (no mask)                         & \best{94.46} & \underline{94.11} & 93.04 & 93.10 & 92.36 & 93.91 & 92.03 & 92.20 & 93.88 \\
OCRBench                               & 80.00 & 79.30 & \best{84.90} & 80.00 & 80.20 & \underline{82.50} & 76.90 & 78.00 & 81.50 \\
ChartQA                                & 86.80 & 86.24 & 85.12 & \underline{86.88} & \best{87.12} & 86.64 & 71.16 & 71.52 & 72.84 \\
MMMU(Val)                              & 55.22 & 52.22 & 51.44 & \underline{56.89} & \best{57.00} & 55.44 & 48.11 & 49.44 & 48.44 \\
MMMU-Pro (Standard)                    & 37.40 & 36.76 & 36.47 & \underline{40.17} & \best{41.50} & 37.11 & 35.49 & 35.43 & 36.19 \\
MMStar                                 & 64.96 & 66.15 & 60.45 & \underline{66.16} & 63.50 & \best{68.01} & 58.24 & 59.25 & 50.73 \\
VStar-Bench                            & 80.10 & \underline{81.68} & \best{84.29} & 69.63 & 70.68 & 78.01 & 75.39 & 75.92 & 73.30 \\
MMBench-EN (Dev)                       & 85.31 & 84.88 & 84.79 & 83.16 & 82.99 & \underline{85.48} & 80.58 & 81.01 & \best{86.34} \\
MMBench-EN (Test)                      & \underline{84.87} & \best{85.82} & 83.91 & 81.28 & 80.21 & 84.30 & 80.16 & 79.65 & 84.19 \\
MME-RealWorld (EN)                     & \best{63.70} & 63.04 & \underline{63.29} & 62.54 & 56.39 & 61.57 & 57.47 & 59.55 & 60.94 \\
MME-RealWorld (CN)                     & 57.99 & \underline{62.26} & \best{64.53} & 60.57 & 52.75 & 54.64 & 44.18 & 46.68 & 55.37 \\
DocVQA(Val)                            & 94.94 & 94.56 & \underline{95.64} & 91.20 & 92.59 & \best{97.85} & 86.51 & 87.05 & 94.94 \\
InfoVQA(Val)                           & 79.37 & 78.11 & \best{83.13} & 74.71 & 75.16 & 79.03 & \underline{83.04} & 82.92 & 72.14 \\
SEED-Bench                             & 73.84 & 72.61 & \best{75.06} & \underline{74.41} & 73.90 & 72.79 & 73.06 & 72.70 & 74.00 \\
SEED-Bench-2-Plus                      & 70.44 & 69.83 & \best{71.19} & \underline{71.15} & 70.44 & 68.99 & 69.35 & 69.39 & 69.26 \\
RealWorldQA                            & \best{71.50} & \underline{70.98} & 69.02 & 64.71 & 63.66 & 67.71 & 67.19 & 66.01 & 69.28 \\
\rowcolor[HTML]{F2F2F2} \best{Average} & \underline{74.50} & 74.33 & \best{74.71} & 72.87 & 71.84 & 74.03 & 69.44 & 69.90 & 70.99 \\
\midrule

\rowcolor[HTML]{EBF2FF} \multicolumn{10}{l}{\textbf{Math \& Reasoning}} \\
MathVision (Test)                      & 31.32 & \best{34.64} & 30.56 & 27.11 & 22.76 & 24.11 & \underline{32.99} & 29.77 & 23.72 \\
MathVision (Mini)                      & 28.95 & \best{37.50} & 29.61 & 30.26 & 24.01 & 26.64 & 33.88 & \underline{36.51} & 18.75 \\
MathVerse (Mini)                       & 54.77 & \best{58.73} & \underline{57.28} & 48.38 & 39.97 & 43.88 & 54.14 & 50.28 & 42.92 \\
MathVista (Mini)                       & \underline{74.30} & \best{75.30} & 73.30 & 72.40 & 68.40 & 63.30 & 65.40 & 65.90 & 61.10 \\
WeMath                         & 65.00 & \underline{70.86} & 66.03 & 61.21 & 50.34 & 55.63 & \best{73.79} & 67.01 & 53.97 \\
\rowcolor[HTML]{F2F2F2} \best{Average} & 50.87 & \best{55.41} & 51.36 & 47.87 & 41.10 & 42.71 & \underline{52.04} & 49.89 & 40.09 \\
\midrule

\rowcolor[HTML]{EBF2FF} \multicolumn{10}{l}{\textbf{Science}} \\
ScienceQA                              & 91.49 & 91.65 & 92.95 & \best{96.09} & \underline{96.04} & 94.86 & 88.89 & 89.15 & 95.07 \\
RxnBench (EN)                      & 82.36 & 79.80 & \best{83.41} & 78.43 & 80.79 & 81.64 & 80.00 & 79.61 & \underline{82.49} \\
RxnBench (ZH)                      & 82.75 & 81.11 & \best{83.61} & 80.26 & 82.16 & 81.70 & 80.26 & 80.59 & \underline{83.34} \\
MolParse                               & \underline{64.90} & \best{65.10} & 4.75 & 5.00 & 6.10 & 2.95 & 4.55 & 5.55 & 3.80 \\
OpenRxn                                & \underline{57.05} & \best{61.75} & 16.15 & 6.55 & 12.35 & 8.00 & 8.15 & 7.35 & 9.15 \\
EMVista                                & 45.95 & \best{50.05} & 44.86 & 46.32 & 44.92 & 43.89 & 39.62 & 42.05 & \underline{49.46} \\
SuperChem (EN) & 12.00 & 11.40 & 10.20 & \underline{12.60} & \best{13.20} & 8.60 & 10.40 & 8.00 & 11.40 \\
SuperChem (CN) & 10.20 & 9.40 & 10.40 & \underline{11.40} & \best{13.60} & 7.60 & 10.00 & 8.60 & \underline{11.40} \\
SmolInstruct & 22.72 & 24.89 & 20.44 & \underline{62.41} & \best{63.54} & 12.21 & 20.27 & 18.63 & 11.99 \\
ProteinLMBench                         & 62.39 & 60.06 & 59.32 & \underline{62.82} & \best{64.41} & 61.55 & 58.58 & 58.90 & 61.97 \\
SFE (EN)                               & \underline{35.90} & \best{36.49} & 28.71 & 33.39 & 35.50 & 25.66 & 29.37 & 27.71 & 30.30 \\
SFE (ZH)                               & \best{35.69} & \underline{34.58} & 28.85 & 31.47 & 33.45 & 24.98 & 27.18 & 26.68 & 27.77 \\
MicroVQA                               & \underline{51.06} & 40.79 & 49.23 & 50.67 & 50.00 & 48.17 & 48.75 & \best{51.82} & 29.65 \\
MSEarth-MCQ                             & 52.59 & 52.87 & 48.56 & \underline{56.50} & \best{56.93} & 54.81 & 46.23 & 46.66 & 50.50 \\
XLRS-Bench-lite                        & 44.83 & 46.98 & 49.98 & \underline{50.50} & \best{51.12} & 44.00 & 44.40 & 45.04 & 47.38 \\
\rowcolor[HTML]{F2F2F2} \best{Average} & \best{50.13} & \underline{49.79} & 42.09 & 45.63 & 46.94 & 40.04 & 39.78 & 39.76 & 40.38 \\
\midrule

\rowcolor[HTML]{EBF2FF} \best{Overall} & \underline{61.42} & \best{61.83} & 58.33 & 58.45 & 57.59 & 56.02 & 55.06 & 54.97 & 54.40 \\
\bottomrule
\end{tabular}
}
\label{tab:main_results}
\end{table*}

\section{Evaluation}
\label{sec:evaluation}

\subsection{Comparison Models}
To evaluate \textbf{Innovator-VL}, we compare it against a range of state-of-the-art multimodal models with similar parameter scales (7B–9B), including: \textbf{Qwen3-VL-8B}~\citep{Qwen3-VL}, \textbf{InternVL3.5-8B}~\citep{zhu2025internvl3}, \textbf{Intern-S1-mini (9B)}~\citep{bai2025intern}, \textbf{LLaVA-OneVision (LLaVA-OV) 1.5-8B}~\citep{an2025llava}, \textbf{MiMo-VL-7B}~\citep{coreteam2025mimovltechnicalreport} (SFT and RL versions), and \textbf{MiniCPM-V 4.5 (8B)}~\citep{yao2024minicpm}. These models provide a robust baseline for assessing our model's performance in general perception, mathematical reasoning, and scientific domains.

\subsection{Benchmarks}
To comprehensively evaluate our model \textbf{Innovator-VL}, we select a set of influential and representative benchmarks and systematically categorize them into three capability dimensions: \textbf{general}, \textbf{math \& reasoning}, and \textbf{science} evaluations. Specifically, we use:

\begin{itemize}
    \item \textbf{General Vision:} AI2D~\citep{AI2D}, OCRBench~\citep{OCRBench}, ChartQA~\citep{ChartQA}, MMMU(Val)~\citep{MMMU}, MMMU-Pro (Standard)~\citep{mmlu-pro}, MMStar~\citep{MMstar}, VStar-Bench~\citep{VStar-Bench}, MMBench-EN (Dev/Test)~\citep{MMBench}, MME-RealWorld (EN/CN)~\citep{MME-realworld}, DocVQA(Val)~\citep{DocVQA}, InfoVQA(Val)~\citep{InfoVQA}, SEED-Bench~\citep{SEED-Bench}, SEED-Bench-2-plus~\citep{seed-banch2}, and RealWorldQA~\citep{RealWorldQA}.
    
    \item \textbf{Math \& Reasoning:} MathVision (Test/Mini)~\citep{MathVision}, MathVerse (Mini)~\citep{MathVerse}, MathVista (Mini)~\citep{MathVista}, and WeMath~\citep{qiao2025we}.
    
    \item \textbf{Science:} ScienceQA~\citep{ScienceQA}, RxnBench (EN/ZH)~\citep{li2025rxnbench}, MolParse\footnote{\url{https://huggingface.co/datasets/InnovatorLab/MolParse}}, OpenRxn\footnote{\url{https://huggingface.co/datasets/InnovatorLab/OpenRxn}}, EMVista\footnote{\url{https://huggingface.co/datasets/InnovatorLab/EMVista}}, SuperChem (EN/CN)~\citep{zhao2025superchem}, SmolInstruct~\citep{smolinstruct}, ProteinLMBench~\citep{shen2024fine}, SFE (EN/ZH)~\citep{zhou2025scientists}, MicroVQA~\citep{MicroVQA}, MSEarth-MCQ~\citep{MSEarth}, and XLRS-Bench-lite~\citep{XLRS-Bench-lite}.
\end{itemize}

These benchmarks are primarily \textbf{multimodal} in nature, enabling an objective and thorough comparison between \textbf{Innovator-VL} and state-of-the-art models across diverse scenarios and capability aspects. More details about these datasets can be found in the Appendix~\ref{Details dataset}.

\subsection{Implementation Details}
We run all evaluations using the \texttt{lmms-eval} framework~\citep{zhang2024lmmsevalrealitycheckevaluation}. For Innovator-VL-8B-Instruct and Innovator-VL-8B-Thinking, we use deterministic decoding (temperature $0.0$, top-$p$ $1.0$) across all benchmarks. Additional evaluation details are provided in Appendix~\ref{sec:evaluation_details}.

\subsection{Performance Comparison}
As summarized in Table~\ref{tab:main_results}, we evaluate \textbf{Innovator-VL} across 37 diverse benchmarks spanning general, mathematical, and scientific domains. Overall, \textbf{Innovator-VL-8B-Thinking} achieves a state-of-the-art (SOTA) average score of \textbf{61.83\%}, surpassing all compared models of similar scale. These results highlight the remarkable performance of our model, which delivers excellent general multimodal capabilities while achieving significant breakthroughs in specialized scientific domains.

\subsubsection{General Multimodal Understanding}
As shown in Table~\ref{tab:main_results}, \textbf{Innovator-VL-8B-Instruct} demonstrates excellent general multimodal capabilities. With an average score of \textbf{74.50\%}, it is on par with the state-of-the-art Qwen3-VL-8B (74.71\%) while significantly outperforming other representative models such as InternVL3.5-8B and LLaVA-OV-1.5-8B. It also demonstrates superior scene understanding on \textbf{MME-RealWorld} and achieves the best performance on \textbf{AI2D} and \textbf{RealWorldQA}. These results indicate that Innovator-VL possesses exceptionally robust visual recognition and instruction-following abilities.

\subsubsection{Multimodal Reasoning}
The effectiveness of our reinforcement learning (RL) strategy is most prominently reflected in the \textbf{Math \& Reasoning} category. \textbf{Innovator-VL-8B-Thinking} achieves a state-of-the-art average score of \textbf{55.41\%}, representing a significant \textbf{4.54\%} absolute improvement over \textbf{Innovator-VL-8B-Instruct}. Notably, \textbf{Innovator-VL-8B-Thinking} attains the highest performance in this category across all compared models. 
These results demonstrate that RL enhances the model's performance in complex reasoning.

\subsubsection{Scientific Knowledge}
As a specialized scientific multimodal model, \textbf{Innovator-VL} exhibits a dominant position in the \textbf{Science} category, with its versions securing the top two average scores (\textbf{50.13\%} and \textbf{49.79\%}), significantly exceeding all general models. 
The most striking advantage is observed in specialized chemistry tasks. For instance, on the \textbf{OpenRxn} and \textbf{MolParse}  benchmarks, Innovator-VL achieves scores exceeding \textbf{57\%} and \textbf{64\%} respectively, while all other baselines fail to exceed 17\%.
Such a substantial performance gap demonstrates that Innovator-VL has successfully internalized complex scientific knowledge.

\begin{figure}[!t]
    \centering
    \includegraphics[width=\linewidth]{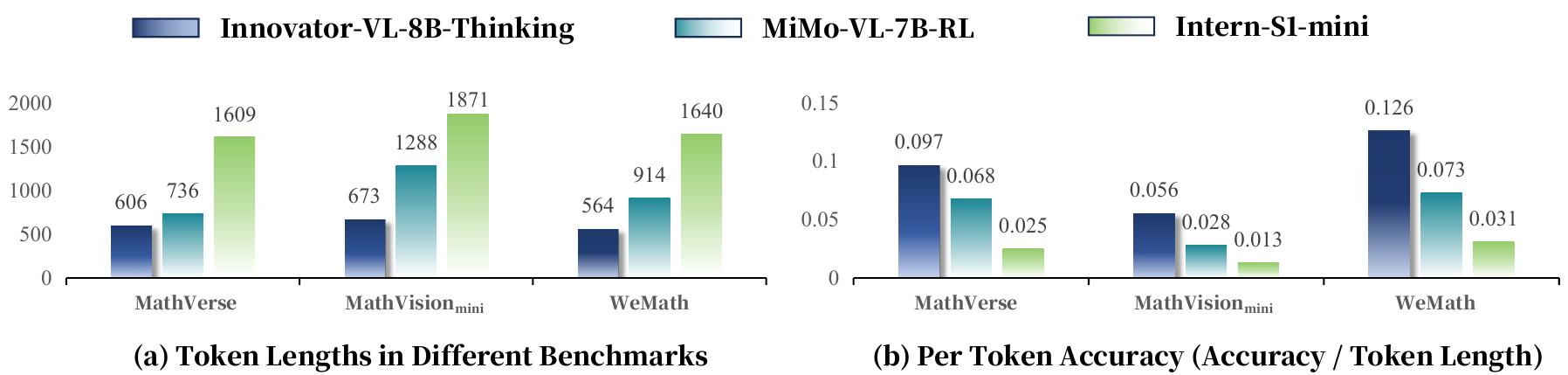}
    \caption{Token efficiency comparison across vision reasoning benchmarks. (a) Average token lengths, showing that Innovator-VL-8B-Thinking generates significantly shorter reasoning chains compared to the other models. (b) Accuracy-to-token ratio, which measures the reasoning efficiency, demonstrating that Innovator-VL-8B-Thinking achieves $1.4\times$ to $2\times$ higher accuracy-to-token ratio than MiMo-VL-7B-RL and $3.9\times$ to $4.3\times$ higher than Intern-S1-mini.}
    \label{fig:token_efficiency}
\end{figure}
\section{On Token Efficiency of Reasoning}
\label{sec:discussion}

Beyond the accuracy improvements demonstrated in Section~\ref{sec:evaluation}, Innovator-VL exhibits remarkable efficiency in reasoning token consumption. As shown in Figure~\ref{fig:token_efficiency} (a), Innovator-VL-8B-Thinking generates significantly shorter reasoning chains compared to Intern-S1-mini~\citep{bai2025intern} and MiMo-VL-7B-RL~\citep{coreteam2025mimovltechnicalreport}. Specifically, our model consumes approximately 62\% to 66\% fewer tokens than Intern-S1-mini and 18\% to 48\% fewer than MiMo-VL across the three benchmarks. On WeMath, for instance, our model requires merely 564 average tokens, whereas Intern-S1-mini needs 1,640 tokens.

More importantly, as illustrated in Figure~\ref{fig:token_efficiency} (b), Innovator-VL-8B-Thinking achieves substantially higher accuracy-to-token ratios across all benchmarks, indicating superior reasoning efficiency. Our model delivers approximately $1.4\times$ to $2\times$ higher accuracy-to-token ratio than MiMo-VL-7B-RL and $3.9\times$ to $4.3\times$ higher than Intern-S1-mini. This metric directly captures the effectiveness of each token consumed during reasoning: higher values indicate that the model extracts more useful information per token, rather than generating redundant or irrelevant content.

This efficiency advantage stems from our reinforcement learning stage that explicitly optimizes for both correctness and conciseness. During RL training, the model learns to identify critical reasoning steps and bypass redundant computations, forming compact yet effective reasoning pathways. This capability is particularly valuable for scientific applications where lengthy inference chains not only incur computational costs but also risk accumulating errors. Compared to recent trends that pursue performance through extensive test-time computation scaling, which often incurs prohibitive token costs, Innovator-VL demonstrates that superior accuracy-efficiency trade-offs can be achieved through principled post-training recipes. Such efficient reasoning ability makes our model more practical for real-world deployment, especially in resource-constrained research environments~\citep{chen2025ipcv} and latency-sensitive applications~\citep{xiong2025prune2drive,yang2025efficientvla,he2025audiomarathon}.

\section{Conclusion and Future Works}
\label{sec:conclusion}

In this work, we present Innovator-VL, a scientific multimodal large language model that achieves strong performance across diverse scientific domains while maintaining excellent general vision capabilities through a fully transparent and reproducible training pipeline. Our work demonstrates three key principles for building scientific multimodal intelligence: first, that competitive scientific reasoning can be achieved with fewer than five million carefully curated samples, challenging the prevailing reliance on massive domain-specific pretraining; second, that scientific alignment and general-purpose capabilities can coexist within a unified framework without mutual degradation; and third, that principled reinforcement learning training can elicit both accurate and concise reasoning, as evidenced by our model's superior token efficiency compared to peer models. Looking forward, we aim to extend Innovator-VL toward real-world scientific workflows such as automated experimental design and cross-modal hypothesis generation, expand its modalities to encompass video, 3D molecular structures, and time-series data, develop lightweight variants through compression techniques to broaden accessibility, and enhance its reasoning through integration with external scientific tools and knowledge bases. We hope that our transparent methodology and the practical foundation established by Innovator-VL will enable the research community to advance scientific multimodal modeling and contribute meaningfully to scientific discovery and innovation.


\bibliographystyle{unsrtnat}
\bibliography{main}

\clearpage
\appendix

\section{More Results}
\subsection{Qualitative Case Study}
To better illustrate the practical strengths of our approach beyond aggregate benchmark scores, we present a curated set of representative cases sampled from the evaluation benchmarks. For each case, our trained model produces the correct answer, while other baseline models fail under the same evaluation setting. We report the original question (and associated image/figure when applicable), together with the full model responses, to highlight where our model’s ability aligns with the task requirements and where competing methods break down. These examples are not meant to be exhaustive; rather, they provide qualitative evidence of improved robustness and generalization in challenging scenarios.
\subsubsection{General Visual Tasks}
\subsection*{Case 1} 

\begin{QuestionBox}{Question}
\centering
\includegraphics[width=1.0\linewidth]{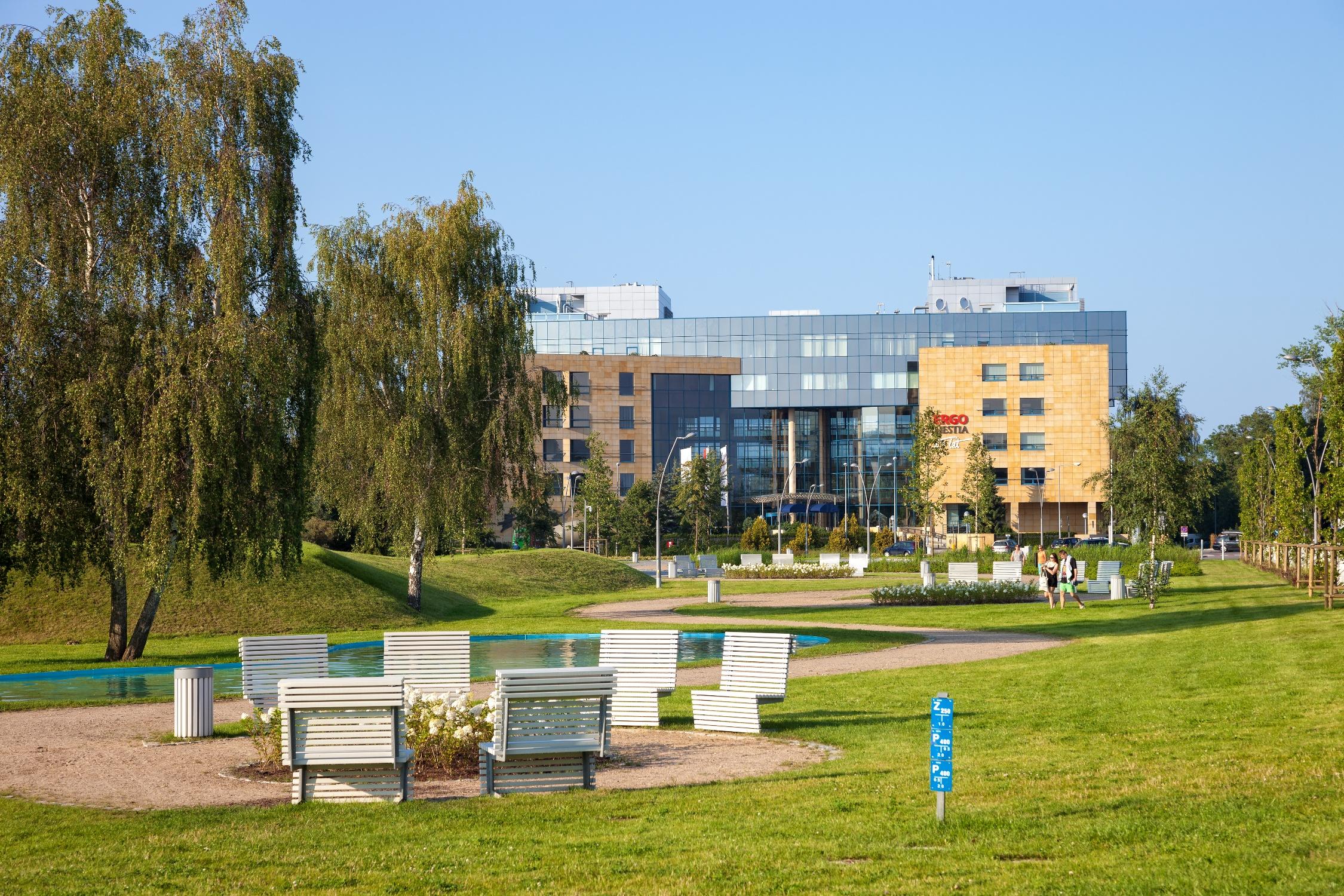}\par 
\vspace{4pt}
\raggedright
What is the color of the handbag?

\medskip
\noindent\textbf{Choices:}
\begin{enumerate}[label=\Alph*.]
  \item brown
  \item black
  \item red
  \item white
\end{enumerate}

Answer with the option's letter from the given choices directly.

\end{QuestionBox}

\begin{AnswerBox}{Answer}
D
\end{AnswerBox}

\subsubsection*{Innovator-VL-8B-Thinking} 

\begin{ThinkingBox}{Model Output}
D
\end{ThinkingBox}

\subsubsection*{Qwen3-VL-8B} 
\begin{ThinkingBox}{Model Output}
B
\end{ThinkingBox}

\subsubsection*{MiMo-VL-7B-RL} 
\begin{ThinkingBox}{Model Output}
B
\end{ThinkingBox}

\subsection*{Case 2} 

\begin{QuestionBox}{Question}
\centering
\includegraphics[width=1.0\linewidth]{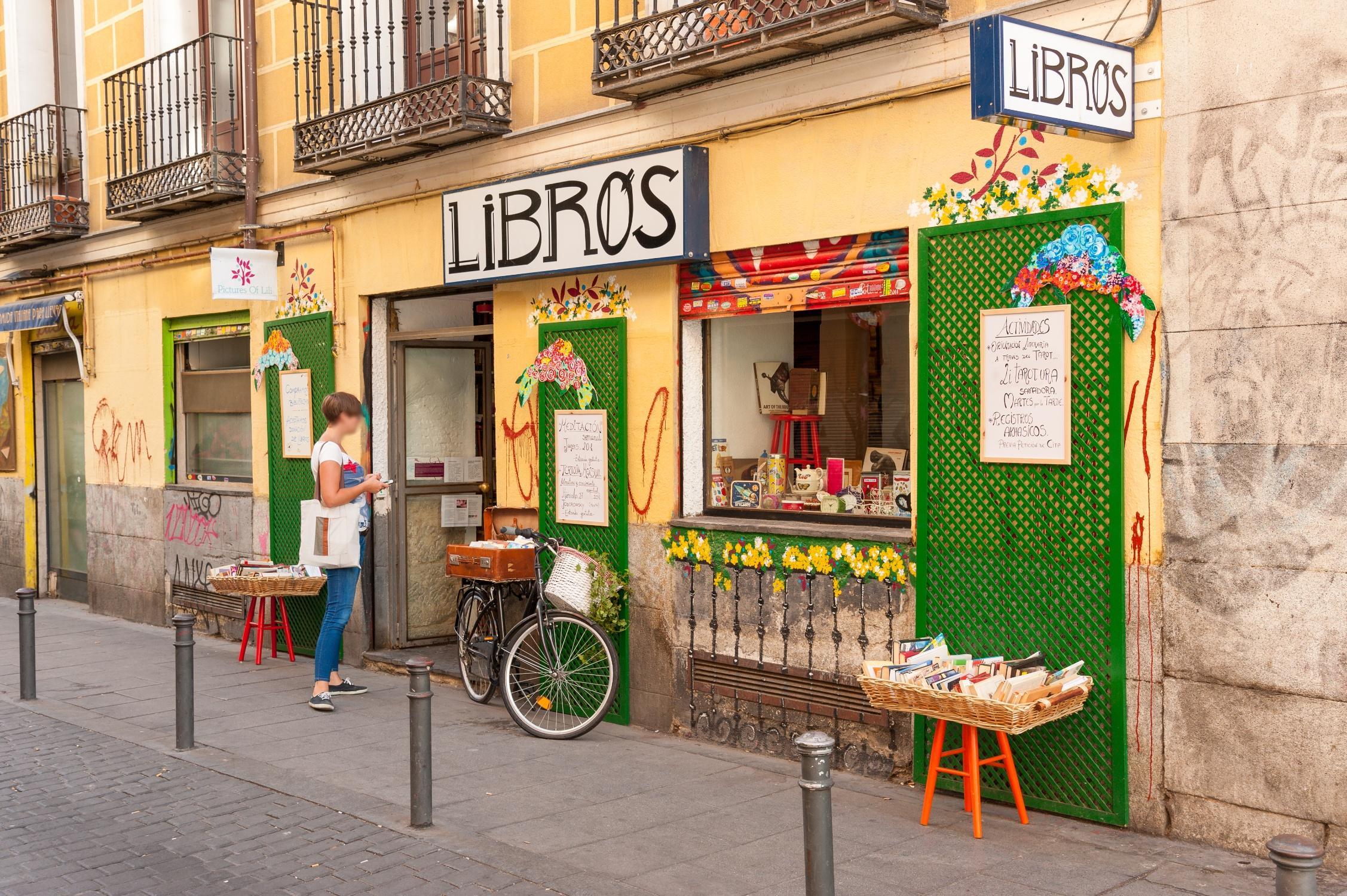}\par 
\vspace{4pt}
\raggedright
What is the color of the clock?

\medskip
\noindent\textbf{Choices:}
\begin{enumerate}[label=\Alph*.]
  \item black
  \item yellow
  \item green
  \item red
\end{enumerate}

Answer with the option's letter from the given choices directly.

\end{QuestionBox}

\begin{AnswerBox}{Answer}
C
\end{AnswerBox}

\subsubsection*{Innovator-VL-8B-Instruct} 

\begin{ThinkingBox}{Model Output}
C
\end{ThinkingBox}

\subsubsection*{InternVL3.5-8B} 
\begin{ThinkingBox}{Model Output}
A
\end{ThinkingBox}

\subsubsection*{Intern-S1-mini (9B)} 
\begin{ThinkingBox}{Model Output}
A
\end{ThinkingBox}

\subsection*{Case 3} 

\begin{QuestionBox}{Question}
\centering
\includegraphics[width=0.6\linewidth]{figs/Appendix-A-More-Results/Case_imgs/realworldqa-rl_0001-402.jpg}\par 
\vspace{4pt}
\raggedright
What direction can we go?

\medskip
\noindent\textbf{Choices:}
\begin{enumerate}[label=\Alph*.]
  \item Straight
  \item Right
  \item Left
\end{enumerate}

Please answer directly with only the letter of the correct option and nothing else.

\end{QuestionBox}

\begin{AnswerBox}{Answer}
C
\end{AnswerBox}

\subsubsection*{Innovator-VL-8B-Thinking} 

\begin{ThinkingBox}{Model Output}
C
\end{ThinkingBox}

\subsubsection*{MiniCPM-V4.5 (8B)} 
\begin{ThinkingBox}{Model Output}
A
\end{ThinkingBox}

\subsubsection*{Qwen3-VL-8B} 
\begin{ThinkingBox}{Model Output}
B
\end{ThinkingBox}

\subsection*{Case 4} 

\begin{QuestionBox}{Question}
\centering
\includegraphics[width=0.8\linewidth]{figs/Appendix-A-More-Results/Case_imgs/realworldqa-sft_0001.jpg}\par 
\vspace{4pt}
\raggedright
The bike rider has a blue backpack.

Please answer directly with a single word or number.

\end{QuestionBox}

\begin{AnswerBox}{Answer}
Yes
\end{AnswerBox}

\subsubsection*{Innovator-VL-8B-Instruct} 

\begin{ThinkingBox}{Model Output}
Yes
\end{ThinkingBox}

\subsubsection*{MiMo-VL-7B-RL} 
\begin{ThinkingBox}{Model Output}
No
\end{ThinkingBox}

\subsubsection*{Intern-S1-mini (9B)} 
\begin{ThinkingBox}{Model Output}
No
\end{ThinkingBox}

\subsubsection{Math \& Reasoning Tasks}
\subsection*{Case 1} 

\begin{QuestionBox}{Question}
\centering
\includegraphics[width=0.2\linewidth]{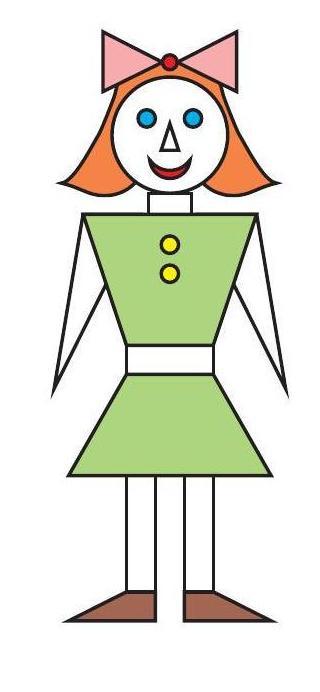}\par 
\vspace{4pt}
\raggedright
How many triangles can you find in the picture?
\end{QuestionBox}

\begin{AnswerBox}{Answer}
5
\end{AnswerBox}

\subsubsection*{Innovator-VL-8B-Thinking} 

\begin{ThinkingBox}{Model Output}
To determine the number of triangles in the given image, let's break it down step by step:

\textbf{Step 1: Analyze the overall structure} \\
The image depicts a simple stick figure with a dress. The dress has a triangular skirt, and there are other triangular shapes in the figure's head and body.

\textbf{Step 2: Identify individual triangles}
\begin{enumerate}
  \item \textbf{Head:}
  \begin{itemize}
    \item The bow on top of the head is composed of two triangles.
    \item The face itself does not contain any triangles, but the eyes and mouth are simple shapes.
  \end{itemize}

  \item \textbf{Dress:}
  \begin{itemize}
    \item The skirt of the dress is a large triangle.
  \end{itemize}

  \item \textbf{Body:}
  \begin{itemize}
    \item There are no additional triangles in the body beyond the dress.
  \end{itemize}

  \item \textbf{Feet:}
  \begin{itemize}
    \item Each foot is a triangle.
  \end{itemize}
\end{enumerate}

\textbf{Step 3: Count the triangles}
\begin{itemize}
  \item \textbf{Bow}: 2 triangles (one for each half of the bow).
  \item \textbf{Skirt}: 1 triangle.
  \item \textbf{Feet}: 2 triangles (one for each foot).
\end{itemize}

\textbf{Step 4: Sum up all the triangles}
\[
\text{Total} = 2 + 1 + 2 = 5.
\]

\textbf{Final Answer:}
\[
\boxed{5}
\]

\end{ThinkingBox}

\subsubsection*{MiniCPM-V4.5 (8B)} 
\begin{ThinkingBox}{Model Output}
Let's analyze the image to find all the triangles:

\begin{enumerate}
  \item First, look at the girl's bow on her head. The bow is a triangle shape.
  \item Next, consider the dress. The main part of the dress (the green section without the belt) has a triangular lower part that forms an upside-down triangle.
\end{enumerate}

Now, let's count each distinct triangle:

\begin{itemize}
  \item The pink bow on top of her head is one triangle.
  \item The bottom part of the green dress (the skirt area) is another triangle.
\end{itemize}

Are there any other triangles? Let's check each part again:
\begin{itemize}
  \item Her face features: eyes and mouth are circles and lines, not triangles.
  \item Arms and legs are straight lines (rectangles or trapezoids).
  \item Belt is a rectangle.
  \item Hair is curved but does not form a triangle in this stylized drawing.
\end{itemize}

So, only the bow and the bottom part of the dress are triangles.

\end{ThinkingBox}

\subsubsection*{InternVL3.5-8B} 
\begin{ThinkingBox}{Model Output}
So, let's count the triangles in the picture step by step. First, I'll look at the obvious large triangles and then check for smaller ones.

\begin{enumerate}
  \item \textbf{The bow on the head:} There are two large triangles forming the bow.
  \item \textbf{The dress:} The dress has a large triangle shape on the top part.
  \item \textbf{The skirt:} The skirt has a large triangle shape.
  \item \textbf{The arms:} Each arm is a triangle, and there are two arms.
  \item \textbf{The legs:} Each leg is a triangle, and there are two legs.
  \item \textbf{The feet:} Each foot is a triangle, and there are two feet.
  \item \textbf{The buttons:} The buttons on the dress are circles, so they don't count as triangles.
\end{enumerate}

Wait, let me recount to make sure I didn't miss any:

\begin{itemize}
  \item Bow: 2 triangles
  \item Dress top: 1 triangle
  \item Skirt: 1 triangle
  \item Arms: 2 triangles
  \item Legs: 2 triangles
  \item Feet: 2 triangles
\end{itemize}

Adding them up:
\[
2 + 1 + 1 + 2 + 2 + 2 = 10.
\]

Hmm, maybe I missed some smaller triangles. Let's check again:

\begin{itemize}
  \item The bow has 2 large triangles.
  \item The dress top is 1 large triangle.
  \item The skirt is 1 large triangle.
  \item Each arm is 1 triangle, so 2 triangles.
  \item Each leg is 1 triangle, so 2 triangles.
  \item Each foot is 1 triangle, so 2 triangles.
\end{itemize}

That still totals 10. But wait, maybe there are more small triangles in the design. Let's see:

\begin{itemize}
  \item The bow might have smaller triangles inside, but they seem to be part of the larger shape.
  \item The dress might have some hidden triangles, but I don't see any.
\end{itemize}

I think 10 is the correct count. Let me confirm once more:

\begin{itemize}
  \item Bow: 2
  \item Dress top: 1
  \item Skirt: 1
  \item Arms: 2
  \item Legs: 2
  \item Feet: 2
\end{itemize}

Total:
\[
2 + 1 + 1 + 2 + 2 + 2 = 10.
\]

So, the total number of triangles should be 10.

\medskip
\textbf{Answer:} \(10\)

\end{ThinkingBox}
\subsection*{Case 2} 

\begin{QuestionBox}{Question}
\centering
\includegraphics[width=0.5\linewidth]{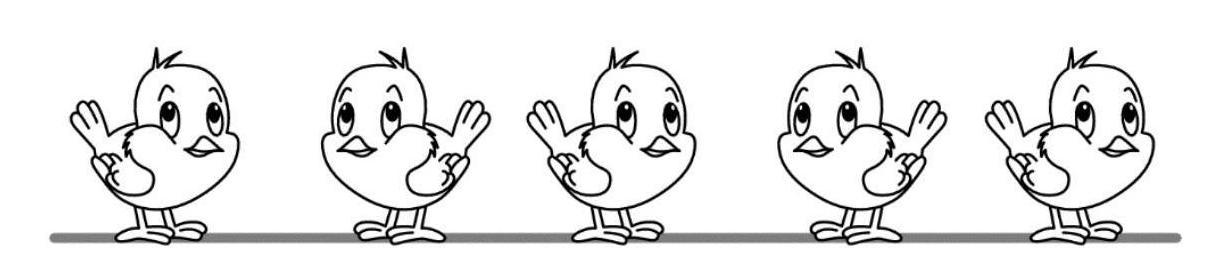}\par 
\vspace{4pt}
\raggedright
Five sparrows are sitting on a rope (see picture). Some of them are looking to the left, some of them are looking to the right. Every sparrow whistles as many times as the number of sparrows he can see sitting in front of him. For example, the third sparrow whistles exactly twice. How often do all sparrows whistle altogether?
\end{QuestionBox}

\begin{AnswerBox}{Answer}
10
\end{AnswerBox}

\subsubsection*{Innovator-VL-8B-Thinking} 

\begin{ThinkingBox}{Model Output}
To solve this problem, we need to determine how many times each sparrow whistles based on the number of sparrows it can see in front of it. Let's analyze the situation step by step.

\textbf{Step 1: Understand the Problem}
\begin{itemize}
  \item There are 5 sparrows sitting on a rope.
  \item Each sparrow whistles as many times as the number of sparrows it can see in front of it.
  \item We need to calculate the total number of whistles from all sparrows.
\end{itemize}

\textbf{Step 2: Analyze Each Sparrow} \\
Let's label the sparrows from left to right as \(S_1, S_2, S_3, S_4,\) and \(S_5\).

\begin{itemize}
  \item \textbf{Sparrow \(S_1\):} It can see 0 sparrows in front of it, so it whistles \(0\) times.
  \item \textbf{Sparrow \(S_2\):} It can see 1 sparrow (\(S_1\)) in front of it, so it whistles \(1\) time.
  \item \textbf{Sparrow \(S_3\):} It can see 2 sparrows (\(S_1\) and \(S_2\)) in front of it, so it whistles \(2\) times.
  \item \textbf{Sparrow \(S_4\):} It can see 3 sparrows (\(S_1, S_2,\) and \(S_3\)) in front of it, so it whistles \(3\) times.
  \item \textbf{Sparrow \(S_5\):} It can see 4 sparrows (\(S_1, S_2, S_3,\) and \(S_4\)) in front of it, so it whistles \(4\) times.
\end{itemize}

\textbf{Step 3: Calculate the Total Number of Whistles}
\[
S_1: 0 \quad S_2: 1 \quad S_3: 2 \quad S_4: 3 \quad S_5: 4
\]
\[
\text{Total whistles} = 0 + 1 + 2 + 3 + 4 = 10.
\]

\[
\boxed{10}
\]

\end{ThinkingBox}

\subsubsection*{Qwen3-VL-8B} 
\begin{ThinkingBox}{Model Output}
Let's analyze the problem step by step.

We are given five sparrows sitting on a rope, and each sparrow whistles as many times as the number of sparrows it can see sitting \textit{in front of him}. The key is the interpretation of ``in front of him'', which depends on the direction each sparrow is facing.

Looking at the image:

\begin{itemize}
  \item The first sparrow (leftmost) is waving with its left hand, suggesting it is facing right.
  \item The second sparrow is waving with its right hand, suggesting it is facing left.
  \item The third sparrow is waving with its left hand, suggesting it is facing right.
  \item The fourth sparrow is waving with its right hand, suggesting it is facing left.
  \item The fifth sparrow (rightmost) is waving with its left hand, suggesting it is facing right.
\end{itemize}

So the sparrows alternate in direction:
\[
\text{right},\ \text{left},\ \text{right},\ \text{left},\ \text{right}.
\]
Odd-numbered sparrows (\(S_1, S_3, S_5\)) face right; even-numbered sparrows (\(S_2, S_4\)) face left.

Now, for a sparrow facing right, ``in front'' means to the right; for a sparrow facing left, ``in front'' means to the left. Therefore:

\begin{itemize}
  \item Sparrow \(S_1\) (facing right) sees \(S_2, S_3, S_4, S_5\): \(4\) sparrows \(\Rightarrow\) whistles \(4\) times.
  \item Sparrow \(S_2\) (facing left) sees \(S_1\): \(1\) sparrow \(\Rightarrow\) whistles \(1\) time.
  \item Sparrow \(S_3\) (facing right) sees \(S_4, S_5\): \(2\) sparrows \(\Rightarrow\) whistles \(2\) times.
  \item Sparrow \(S_4\) (facing left) sees \(S_3\): \(1\) sparrow \(\Rightarrow\) whistles \(1\) time.
  \item Sparrow \(S_5\) (facing right) sees none: \(0\) sparrows \(\Rightarrow\) whistles \(0\) times.
\end{itemize}

The statement ``For example, the third sparrow whistles exactly twice.'' matches this calculation for \(S_3\).

Thus the total number of whistles is:
\[
4 + 1 + 2 + 1 + 0 = 8.
\]

\medskip
\textbf{Answer:} \(8\)

\end{ThinkingBox}

\subsubsection*{MiMo-VL-7B-RL} 
\begin{ThinkingBox}{Model Output}
Let's analyze this step-by-step:

\begin{enumerate}
  \item \textbf{Identify which sparrows are looking left and which are looking right.} \\
  From the image:
  \begin{itemize}
    \item Sparrows \(1\) and \(3\) are looking left.
    \item Sparrows \(2\), \(4\), and \(5\) are looking right.
  \end{itemize}

  \item \textbf{Calculate how many times each sparrow whistles.}
  \begin{itemize}
    \item Sparrow \(1\) (looking left): can see 2 sparrows in front of him (\(2\) and \(3\)) \(\Rightarrow\) whistles 2 times.
    \item Sparrow \(2\) (looking right): can see 3 sparrows in front of him (\(3\), \(4\), and \(5\)) \(\Rightarrow\) whistles 3 times.
    \item Sparrow \(3\) (looking left): can see 2 sparrows in front of him (\(4\) and \(5\)) \(\Rightarrow\) whistles 2 times.
    \item Sparrow \(4\) (looking right): can see 1 sparrow in front of him (\(5\)) \(\Rightarrow\) whistles 1 time.
    \item Sparrow \(5\) (looking right): can see 0 sparrows in front of him \(\Rightarrow\) whistles 0 times.
  \end{itemize}

  \item \textbf{Sum up the whistles.}
  \[
  2 + 3 + 2 + 1 + 0 = 8.
  \]
\end{enumerate}

Therefore, all sparrows whistle a total of 8 times.

\end{ThinkingBox}
\subsubsection{Scientific Tasks}
\subsection*{Case 1} 

\begin{QuestionBox}{Question}
\centering
\includegraphics[width=0.5\linewidth]{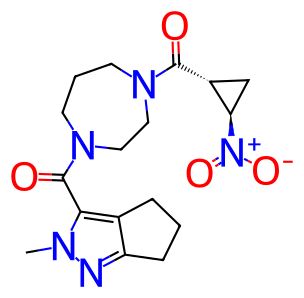}\par 
\vspace{4pt}
\raggedright
\textbf{Please write the Extended SMILES (E-SMILES) of the given molecule image.}

\vspace{6pt}
\noindent\textbf{Below is the definition of Extended SMILES (E-SMILES) Format Writing Rules:}

\medskip
The Extended SMILES (E-SMILES) format used in the MolParser paper is designed to augment a standard RDKit-compatible SMILES string with additional machine-readable annotations, especially useful for representing Markush structures, abstract rings, repeating groups, and other extensions commonly seen in chemical patents.
The general form of an E-SMILES string is:
\begin{center}
\texttt{SMILES\textless sep\textgreater EXTENSION}
\end{center}

\noindent\textbf{Components:}
\begin{enumerate}
  \item \textbf{SMILES}\\
  This is a valid RDKit-compatible SMILES string.\\
  To ensure consistency, the atom indexing starts from 0. You can generate this using RDKit with \texttt{rootedAtAtom=0}.

  \item \textbf{\texttt{\textless sep\textgreater}}\\
  A special delimiter separating the base SMILES string from the extension part.

  \item \textbf{EXTENSION}\\
  The extension part contains structured XML-like tokens providing additional information.\\
  These tokens annotate atom positions, ring positions, and other special features.\\
  There are three types of tags:
  \begin{itemize}
    \item \texttt{\textless a\textgreater ATOM\_INDEX:[GROUP\_NAME]\textless/a\textgreater} --- attaches a group to a specific atom.
    \item \texttt{\textless r\textgreater RING\_INDEX:[GROUP\_NAME]\textless/r\textgreater} --- attaches a group to any position on a ring.
    \item \texttt{\textless c\textgreater CIRCLE\_INDEX:[CIRCLE\_NAME]\textless/c\textgreater} --- defines abstract repeating rings.
  \end{itemize}
  A special token \texttt{\textless dum\textgreater} can be used to represent a connection point.
\end{enumerate}

\noindent\textbf{Definitions:}
\begin{description}
  \item[ATOM\_INDEX:] Index of the atom in the SMILES string (starting from 0).
  \item[RING\_INDEX:] Index of the ring in the molecule (starting from 0).
  \item[CIRCLE\_INDEX:] Index of the abstract repeating ring (starting from 0).
  \item[GROUP\_NAME:] The name of the substituent or group, which may be:
  \begin{itemize}
    \item A standard group: \texttt{R}, \texttt{X}, \texttt{Y}, \texttt{Z}, \texttt{Ph}, \texttt{Me}, \texttt{OMe}, \texttt{CF3}, etc.
    \item An arbitrary or custom-defined group with superscripts or subscripts, e.g., \texttt{R[1]}, \texttt{X[2]}, \texttt{R1}, \texttt{R\textsubscript{3}}.
  \end{itemize}
\end{description}

\noindent\textbf{Examples:}
\begin{enumerate}
  \item \texttt{*c1ccccc1\textless sep\textgreater\textless a\textgreater 0:R[1]\textless/a\textgreater}\\
  $\rightarrow$ Adds substituent \texttt{R[1]} to atom 0.

  \item \texttt{clccecc1\textless sep\textgreater\textless r\textgreater 0:R[1]\textless/r\textgreater\textless r\textgreater 0:R[2]\textless/r\textgreater}\\
  $\rightarrow$ Adds two substituents \texttt{R[1]}, \texttt{R[2]} to ring 0.

  \item \texttt{*C1*C(*)C(C(*)C2=CC=N=C2)N=1\textless sep\textgreater\textless a\textgreater 0:R[4]\textless/a\textgreater\textless a\textgreater 1:X\textless/a\textgreater\textless r\textgreater 1:R[5]\textless/r\textgreater}\\
  $\rightarrow$ Complex structure with atom-level and ring-level substitutions.

  \item \texttt{*c(NC(*)C(*)C(*)C(*)C(=O)N(*)\textless sep\textgreater}\\
\texttt{\textless a\textgreater 0:R[1]\textless/a\textgreater\textless a\textgreater 2:R[3]\textless/a\textgreater\textless a\textgreater 5:R[2]\textless/a\textgreater}\\
\texttt{\textless a\textgreater 7:R[5]\textless/a\textgreater\textless a\textgreater 8:R[4]\textless/a\textgreater\textless c\textgreater 0:B\textless/c\textgreater}\\
\texttt{\textless a\textgreater 13:R[7]\textless/a\textgreater\textless a\textgreater 14:R[6]\textless/a\textgreater}\\

  $\rightarrow$ Shows the usage of abstract ring \texttt{\textless c\textgreater}, with multiple atom-level substitutions.
\end{enumerate}

\end{QuestionBox}

\begin{AnswerBox}{Answer}
\texttt{Cn1nc2c(c1C(=O)N1CCCN(C(=O)[C@@H]3C[C@H]3[N+](=O)[O-])CC1)CCC2\textless sep\textgreater}

\end{AnswerBox}

\subsubsection*{Innovator-VL-8B-Instruct} 

\begin{ThinkingBox}{Model Output}
\texttt{Cn1nc2c(c1C(=O)N1CCCN(C(=O)[C@@H]3C[C@H]3[N+](=O)[O-])CC1)CCC2\textless sep\textgreater}
\end{ThinkingBox}

\subsubsection*{InternVL3.5-8B} 
\begin{ThinkingBox}{Model Output}
To write the E-SMILES for the given molecule, we need to identify the SMILES string and then add any necessary extensions. Let's break down the structure:

\medskip
\noindent\textbf{SMILES String}\\
The SMILES representation of the molecule is:
\begin{quote}\ttfamily
CN1N=C2CCCC2=C1C(=O)N1CCN(C(=O)[C@@H]2C[C@H]2[N+](=O)[O-])CC1
\end{quote}

\noindent\textbf{Extensions}\\
In this molecule, there are no explicit substituents or abstract rings that require additional annotations beyond the standard SMILES. Therefore, the E-SMILES will only include the SMILES string.

\medskip
\noindent\textbf{E-SMILES}
\begin{quote}\ttfamily
CN1N=C2CCCC2=C1C(=O)N1CCN(C(=O)[C@@H]2C[C@H]2[N+](=O)[O-])CC1\textless sep\textgreater
\end{quote}

This E-SMILES string follows the format by providing the SMILES and an empty extension part, as no further annotations are needed for this structure.
\end{ThinkingBox}

\subsubsection*{Intern-S1-mini (9B)} 
\begin{ThinkingBox}{Model Output}
To write the Extended SMILES (E-SMILES) of the given molecule, we need to follow the rules provided. Here's a step-by-step breakdown:

\medskip
\begin{enumerate}
  \item \textbf{SMILES Representation:}\\
  First, we need to write the SMILES string for the given molecule. The SMILES string for the molecule is:
  \begin{quote}\ttfamily
  CN1N=C2CCCC2=C1C(=O)N1CCCN(C(=O)[C@@H]2C[C@H]2[N+]([O-])=O)CC1
  \end{quote}

  \item \textbf{Extension Part:}\\
  Next, we need to add the extension part to the SMILES string. The extension part contains structured XML-like tokens providing additional information.

  \begin{itemize}
    \item \textbf{Atom Indexing:} We need to identify the atom indices where we want to attach groups. In this case, we can attach groups to the nitrogen atom in the pyrazole ring (atom index 0) and the nitrogen atom in the piperazine ring (atom index 1).

    \item \textbf{Group Names:} We can use standard group names like \texttt{R}, \texttt{X}, \texttt{Y}, \texttt{Z}, \texttt{Ph}, \texttt{Me}, \texttt{OMe}, \texttt{CF3}, etc. For this example, let's use \texttt{R} for the groups.

    \item \textbf{Delimiter:} We use the special delimiter \texttt{\textless sep\textgreater} to separate the base SMILES
  \end{itemize}
\end{enumerate}

\end{ThinkingBox}

\subsection*{Case 2} 

\begin{QuestionBox}{Question}
\centering
\includegraphics[width=0.5\linewidth]{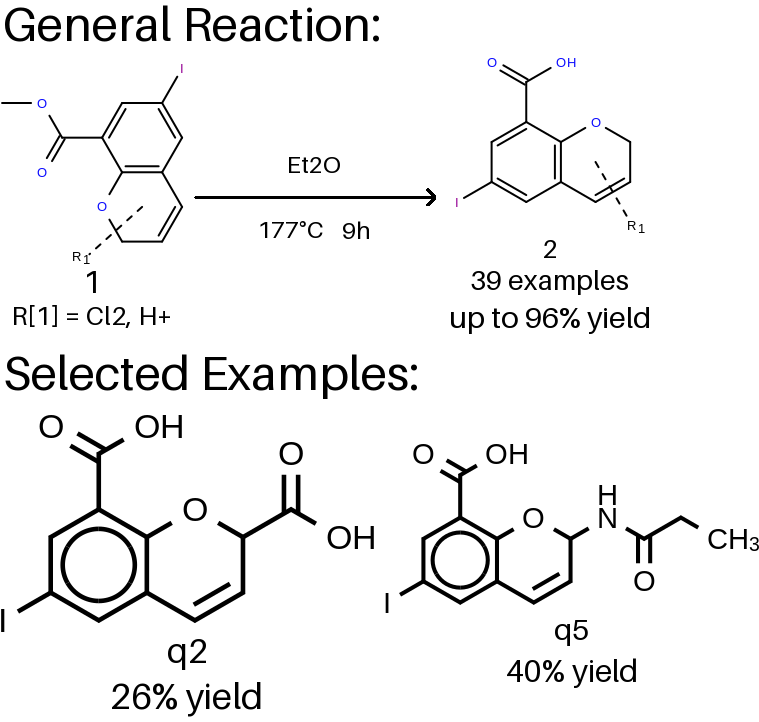}\par 
\vspace{4pt}
\raggedright
Return the ESMILES of the select example with label \texttt{q2}.

\end{QuestionBox}

\begin{AnswerBox}{Answer}
\texttt{\textless E-SMILES\textgreater OC(=O)c1cc(I)cc2c1OC(C(=O)O)C=C2\textless/E-SMILES\textgreater}

\end{AnswerBox}

\subsubsection*{Innovator-VL-8B-Thinking} 

\begin{ThinkingBox}{Model Output}
\texttt{\textless E-SMILES\textgreater OC(=O)c1cc(I)cc2c1OC(C(=O)O)C=C2\textless/E-SMILES\textgreater}
\end{ThinkingBox}

\subsubsection*{Qwen3-VL-8B} 
\begin{ThinkingBox}{Model Output}
\begin{quote}\ttfamily
O=C(O)c1cc2c(cc1I)OC=C2C(=O)O
\end{quote}

\end{ThinkingBox}

\subsubsection*{Intern-S1-mini (9B)} 
\begin{ThinkingBox}{Model Output}
The ESMILES (Extended SMILES) representation of the selected example with label \texttt{q2} is:

\begin{quote}\ttfamily
COC(=O)C1=C(I)C=CC2=C1OC(C(=O)O)C=C2
\end{quote}

This SMILES string represents the structure of the compound shown in the image labeled \texttt{q2}.

\end{ThinkingBox}

\subsection*{Case 3} 

\begin{QuestionBox}{Question}
\centering
\includegraphics[width=0.8\linewidth]{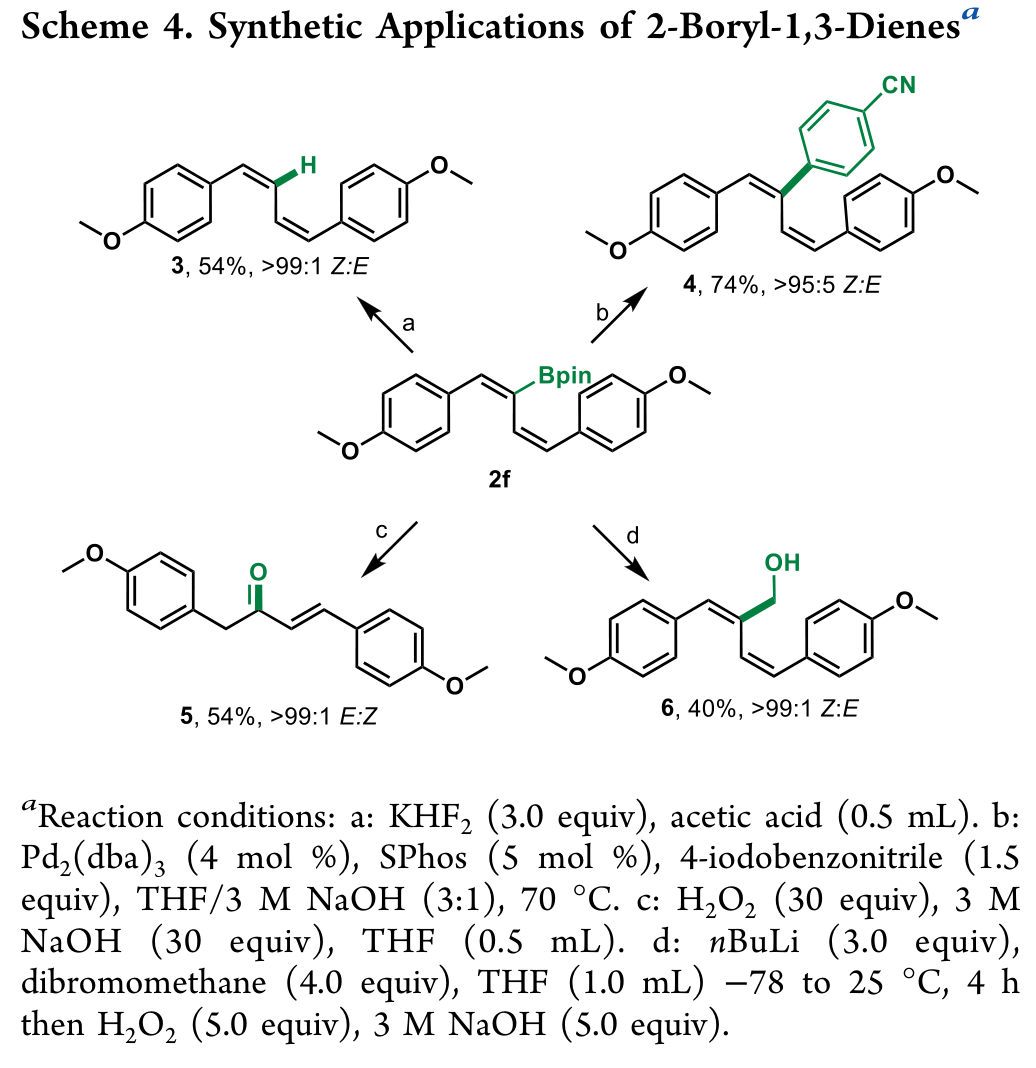}\par 
\vspace{4pt}
\raggedright
Question: Please summarize the four derivatization reactions (a--d) centered on 2-boryl-1,3-diene 2f in Scheme 4 and their corresponding products (3--6).

\medskip
\noindent\textbf{Choices:}
\begin{enumerate}[label=\Alph*.]
  \item Scheme 4 shows four reaction pathways starting from 2-boryl-1,3-diene 2f: \\
  a) KHF$_2$/AcOH effects removal of Bpin to give the deborylated product 3 (yield 54\%, $>$99:1 Z:E); \\
  b) Pd$_2$(dba)$_3$/SPhos-catalyzed Suzuki coupling with 4-iodobenzonitrile affords the arylated product 4 (yield 74\%, $>$95:5 Z:E); \\
  c) H$_2$O$_2$/NaOH oxidation of the boronate gives the $\alpha,\beta,\gamma,\delta$-unsaturated ketone product 5 (yield 54\%, $>$99:1 Z:E); \\
  d) \textit{n}BuLi/CH$_2$Br$_2$ insertion followed by H$_2$O$_2$ oxidation provides the allylic alcohol product 6 (yield 40\%, $>$99:1 Z:E).

  \item Scheme 4 shows four reaction pathways starting from 2-boryl-1,3-diene 2f: \\
  a) KHF$_2$/AcOH effects removal of Bpin to give the deborylated product 3 (yield 54\%, 99:1 Z:E); \\
  b) Pd$_2$(dba)$_3$/SPhos-catalyzed Suzuki coupling with 4-iodobenzonitrile affords the arylated product 4 (yield 74\%, $>$95:5 Z:E); \\
  c) H$_2$O$_2$/NaOH oxidation of the boronate gives the $\alpha,\beta,\gamma,\delta$-unsaturated ketone product 5 (yield 54\%, $>$99:1 E:Z); \\
  d) \textit{n}BuLi/CH$_2$Br$_2$ insertion followed by H$_2$O$_2$ oxidation provides the allylic alcohol product 6 (yield 40\%, $>$99:1 Z:E).

  \item Scheme 4 shows four reaction pathways starting from 2-boryl-1,3-diene 2f: \\
  a) KHF$_2$/AcOH effects removal of Bpin to give the deborylated product 3 (yield 54\%, $>$99:1 Z:E); \\
  b) Pd$_2$(dba)$_3$/SPhos-catalyzed Suzuki coupling with 4-iodobenzonitrile affords the arylated product 4 (yield 74\%, $>$95:5 Z:E); \\
  c) H$_2$O$_2$/NaOH oxidation of the boronate gives the $\alpha,\beta,\gamma,\delta$-unsaturated ketone product 5 (yield 54\%, $>$99:1 E:Z); \\
  d) \textit{n}BuLi/CH$_2$Br$_2$ insertion followed by H$_2$O$_2$ oxidation provides the allylic alcohol product 6 (yield 40\%, $>$99:1 Z:E).

  \item Scheme 4 shows four reaction pathways starting from 2-boryl-1,3-diene 2f: \\
  a) KHF$_2$/AcOH effects removal of Bpin to give the deborylated product 3 (yield 54\%, $>$99:1 Z:E); \\
  b) Pd$_2$(dba)$_3$/SPhos-catalyzed Suzuki coupling with 4-iodobenzonitrile affords the arylated product 4 (yield 74\%, $>$95:5 Z:E); \\
  c) H$_2$O$_2$/NaOH oxidation of the boronate gives the $\alpha,\beta,\gamma,\delta$-unsaturated ketone product 5 (yield 54\%, $>$99:1 E:Z); \\
  d) \textit{n}BuLi/CH$_2$Br$_2$ insertion followed by O$_2$ oxidation provides the allylic alcohol product 6 (yield 40\%, $>$99:1 Z:E).
\end{enumerate}

\medskip
Based on the image and the question, choose the most appropriate answer.

\noindent\textbf{Only output a single letter (A, B, C, or D).} Do NOT output any other text or explanation.

\end{QuestionBox}

\begin{AnswerBox}{Answer}
C
\end{AnswerBox}

\subsubsection*{Innovator-VL-8B-Instruct} 

\begin{ThinkingBox}{Model Output}
C
\end{ThinkingBox}

\subsubsection*{InternVL3.5-8B} 
\begin{ThinkingBox}{Model Output}
B
\end{ThinkingBox}

\subsubsection*{Intern-S1-mini (9B)} 
\begin{ThinkingBox}{Model Output}
A
\end{ThinkingBox}

\subsection*{Case 4} 
\begin{CJK}{UTF8}{gbsn}
\begin{QuestionBox}{Question}
\centering
\includegraphics[width=0.9\linewidth]{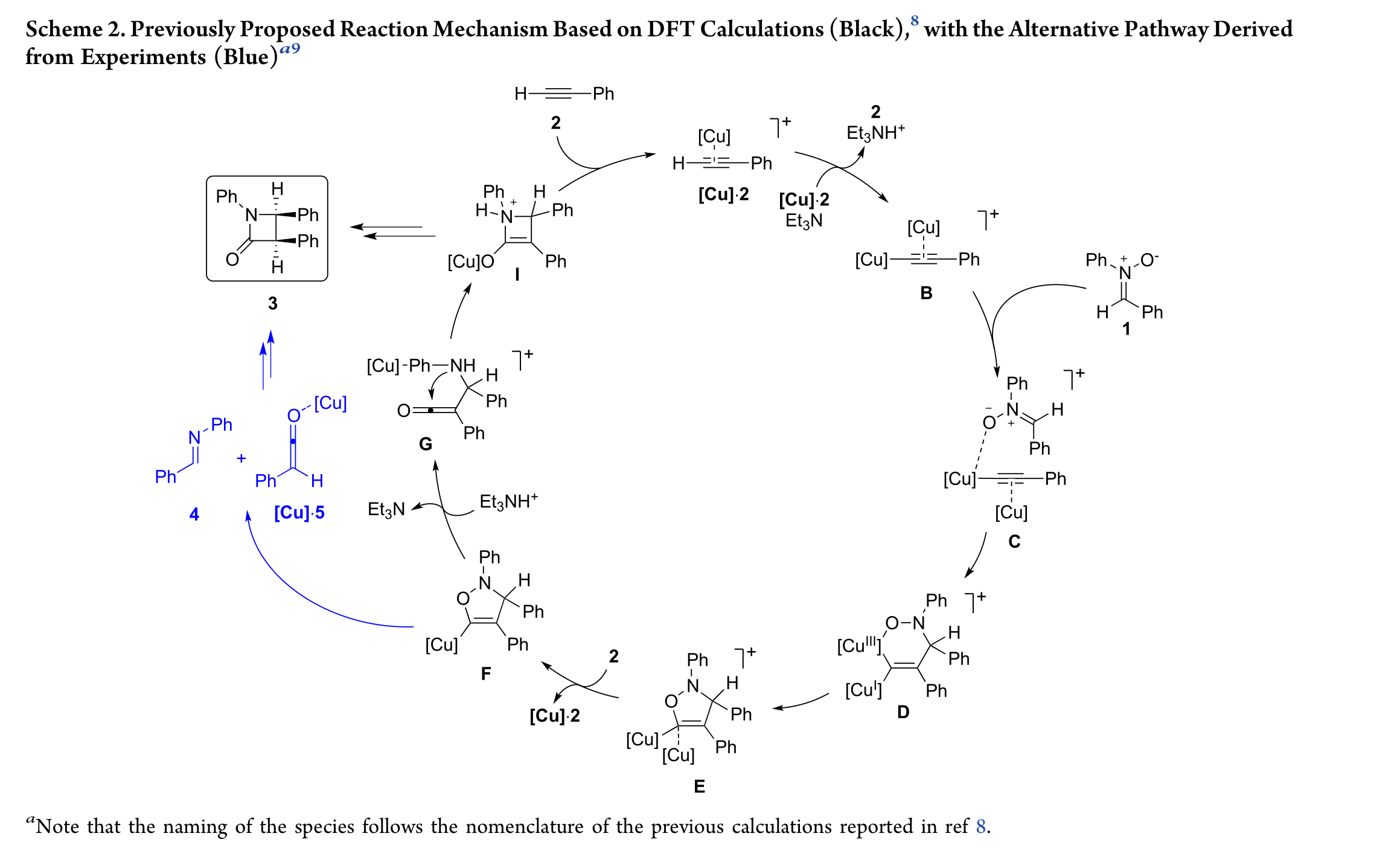}\par 
\vspace{4pt}
\raggedright
问题：该反应的总体转化通式是什么？请用结构简式描述主要底物和产物。

\medskip
\noindent\textbf{选项：}
\begin{enumerate}[label=\Alph*.]
  \item 底物苯基炔（Ph--C$\equiv$CH）与N-氧化亚胺1在Cu催化和Et$_3$N作用下，最终生成五元环化合物3。
  \item 底物苯基炔（Ph--C$\equiv$CH）与N-氧化亚胺1在Cu催化和Et$_3$N作用下，最终生成四元环化合物3。
  \item 底物苯基炔（Ph--C$\equiv$CH）与N-氧化亚胺1在Cu催化和吡啶作用下，最终生成四元环化合物3。
  \item 底物苯基炔（Ph--C$\equiv$CH）与N-氧化亚胺1在Cu催化和Et$_3$N作用下，最终生成桥环化合物3。
\end{enumerate}

\medskip
请根据图像和问题，从以上四个选项中选择最合适的答案。

\noindent\textbf{只输出单个字母（A, B, C 或 D），不要输出选项内容，也不要输出任何解释。}

\end{QuestionBox}

\begin{AnswerBox}{Answer}
B
\end{AnswerBox}

\subsubsection*{Innovator-VL-8B-Thinking} 

\begin{ThinkingBox}{Model Output}
B
\end{ThinkingBox}

\subsubsection*{MiMo-VL-7B-RL} 
\begin{ThinkingBox}{Model Output}
A
\end{ThinkingBox}

\subsubsection*{MiniCPM-V4.5 (8B)} 
\begin{ThinkingBox}{Model Output}
A
\end{ThinkingBox}

\end{CJK}
\subsection*{Case 5} 

\begin{QuestionBox}{Question}
\centering
\includegraphics[width=0.5\linewidth]{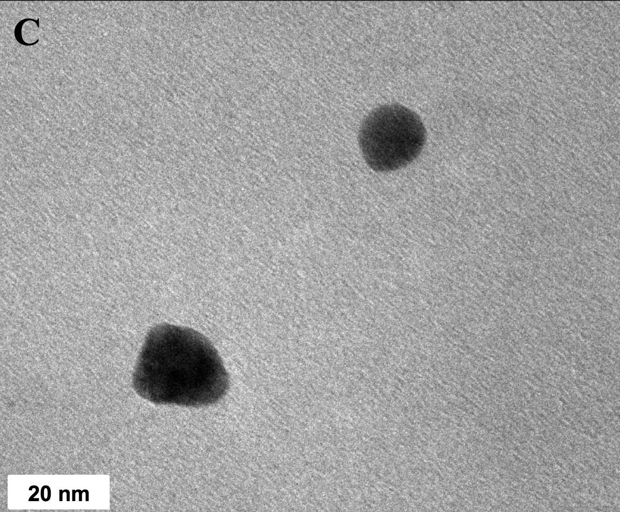}\par 
\vspace{4pt}
\raggedright
Carefully examine the electron microscopy image, select the most accurate diagnosis based on the visual evidence observed in the microstructure, and answer the question: \emph{Is the particle-area distribution uniform?}
(Define non-uniform as $\frac{\text{max/min diameter}}{\text{mean}} > 30\%$) by choosing the most suitable option from:
\begin{itemize}
  \item A. Yes
  \item B. No
  \item C. Uncertain
\end{itemize}

\end{QuestionBox}

\begin{AnswerBox}{Answer}
A
\end{AnswerBox}

\subsubsection*{Innovator-VL-8B-Instruct} 

\begin{ThinkingBox}{Model Output}
A
\end{ThinkingBox}

\subsubsection*{Qwen3-VL-8B} 
\begin{ThinkingBox}{Model Outpu}
B
\end{ThinkingBox}

\subsubsection*{MiMo-VL-7B-RL} 
\begin{ThinkingBox}{Model Output}
B
\end{ThinkingBox}

\subsection*{Case 6} 

\begin{QuestionBox}{Question}
\centering
\includegraphics[width=1\linewidth]{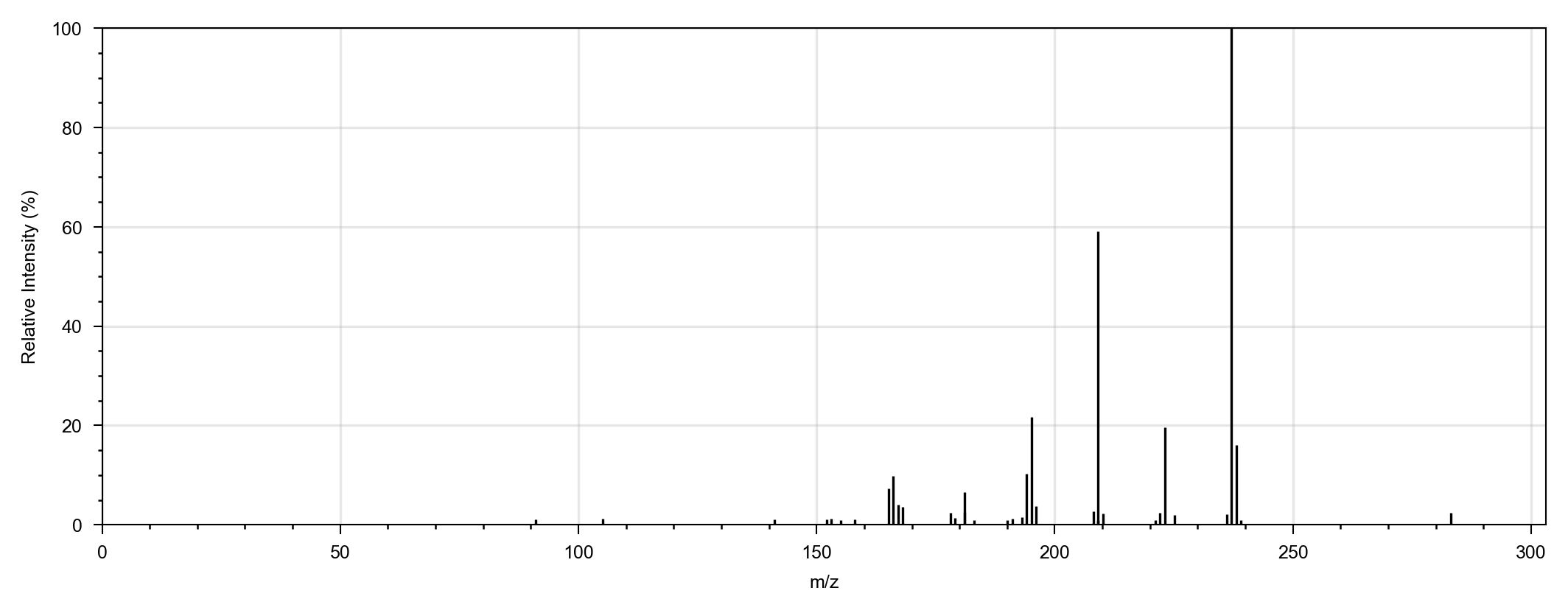}\par 
\vspace{4pt}
\raggedright
You are an expert in life and need to solve the following question.
The question is an exact match question. Answer the question using a single word or phrase.

From the compound's MS/MS spectrum, determine the number of sulfur atoms based on isotope pattern and other compound information in the spectrum.

\end{QuestionBox}

\begin{AnswerBox}{Answer}
0
\end{AnswerBox}

\subsubsection*{Innovator-VL-8B-Thinking} 

\begin{ThinkingBox}{Model Output}
0
\end{ThinkingBox}

\subsubsection*{Qwen3-VL-8B} 
\begin{ThinkingBox}{Model Output}
2
\end{ThinkingBox}

\subsubsection*{MiMo-VL-7B-RL} 
\begin{ThinkingBox}{Model Output}
1
\end{ThinkingBox}

\subsection*{Case 7} 

\begin{QuestionBox}{Question}
\centering
\includegraphics[width=0.8\linewidth]{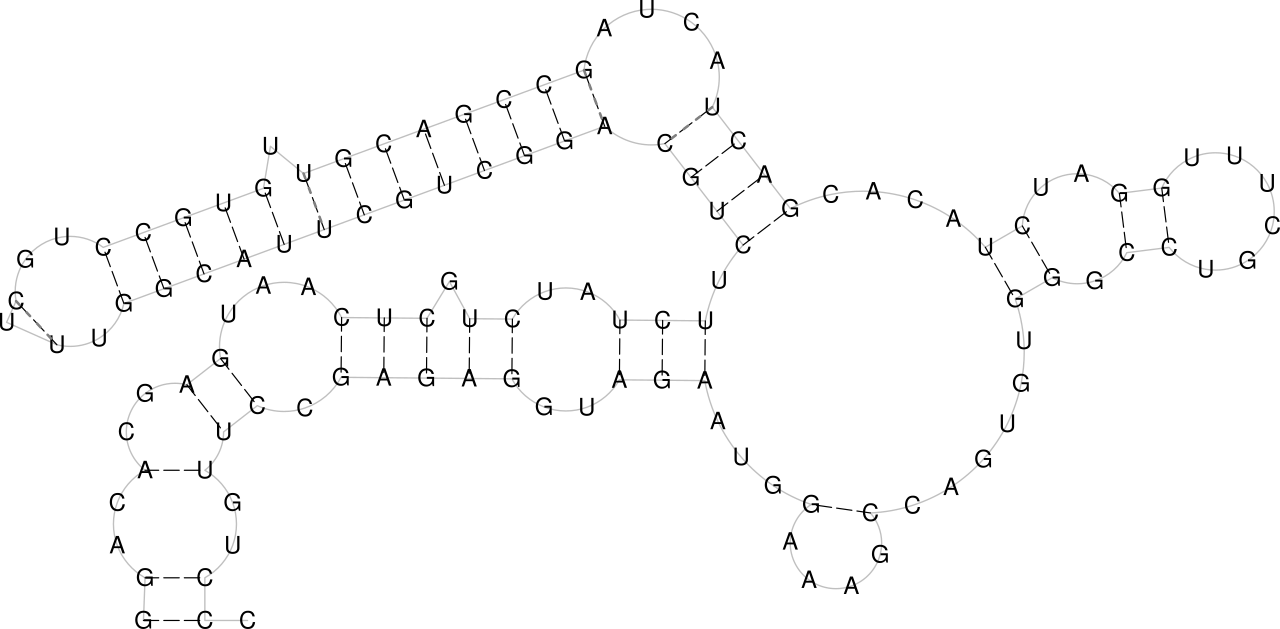}\par 
\vspace{4pt}
\raggedright
You are an expert in life and need to solve the following question.
The question is a multiple-choice question. Answer with the option letter from the given choices.

Based on the RNA secondary structure diagram, which of the following structures might the RNA contain? (Multiple choices)
\begin{enumerate}[label=(\Alph*)]
  \item stem
  \item Hairpin loop
  \item Interior loop
  \item bulge loop
  \item Multi-branched loop
  \item Pseudoknot
\end{enumerate}

\end{QuestionBox}

\begin{AnswerBox}{Answer}
ABCD
\end{AnswerBox}

\subsubsection*{Innovator-VL-8B-Instruct} 

\begin{ThinkingBox}{Model Output}
ABCD
\end{ThinkingBox}

\subsubsection*{Qwen3-VL-8B} 
\begin{ThinkingBox}{Model Output}
ABCDE
\end{ThinkingBox}

\subsubsection*{MiMo-VL-7B-RL} 
\begin{ThinkingBox}{Model Output}
ABCDEF
\end{ThinkingBox}

\subsection*{Case 8} 

\begin{QuestionBox}{Question}
\centering
\includegraphics[width=0.5\linewidth]{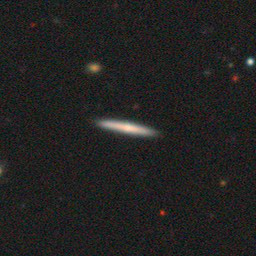}\par 
\vspace{4pt}
\raggedright
You are an expert in astronomy and need to solve the following question.
The question is a multiple-choice question. Answer with the option letter from the given choices.
What is the structure of galaxy in the image?
\begin{enumerate}[label=(\Alph*)]
  \item Unbarred Tight Spiral Galaxies
  \item Unbarred Loose Spiral Galaxies
  \item Cigar Shaped Smooth Galaxies
  \item Edge-on Galaxies with out Bulge
\end{enumerate}

\end{QuestionBox}

\begin{AnswerBox}{Answer}
D
\end{AnswerBox}

\subsubsection*{Innovator-VL-8B-Instruct} 

\begin{ThinkingBox}{Model Output}
D
\end{ThinkingBox}

\subsubsection*{Qwen3-VL-8B} 
\begin{ThinkingBox}{Model Output}
C
\end{ThinkingBox}

\subsubsection*{MiMo-VL-7B-RL} 
\begin{ThinkingBox}{Model Output}
C
\end{ThinkingBox}

\subsection*{Case 9} 
\begin{CJK}{UTF8}{gbsn}
\begin{QuestionBox}{Question}
\centering
\includegraphics[width=0.5\linewidth]{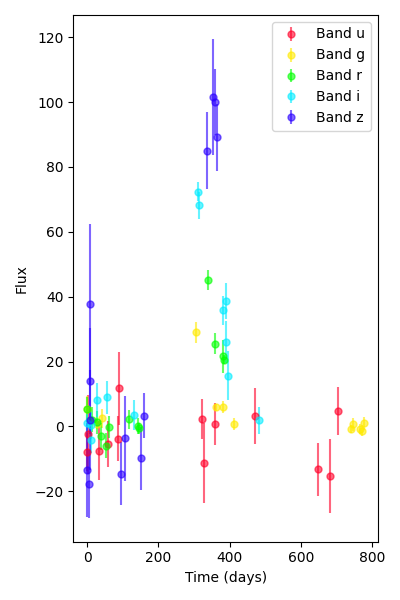}\par 
\vspace{4pt}
\raggedright
根据所提供的不同波段的图像，这是什么类型的光变曲线？
\begin{enumerate}[label=(\Alph*)]
  \item EB
  \item KN
  \item SNII
  \item SNIbc
\end{enumerate}

\end{QuestionBox}

\begin{AnswerBox}{Answer}
C
\end{AnswerBox}

\subsubsection*{Innovator-VL-8B-Thinking} 

\begin{ThinkingBox}{Model Output}
C
\end{ThinkingBox}

\subsubsection*{InternVL3.5-8B} 
\begin{ThinkingBox}{Model Output}
B
\end{ThinkingBox}

\subsubsection*{Intern-S1-mini (9B)} 
\begin{ThinkingBox}{Model Output}
D
\end{ThinkingBox}

\end{CJK}
\subsection*{Case 10} 
\begin{CJK}{UTF8}{gbsn}
\begin{QuestionBox}{Question}
\centering
\includegraphics[width=0.5\linewidth]{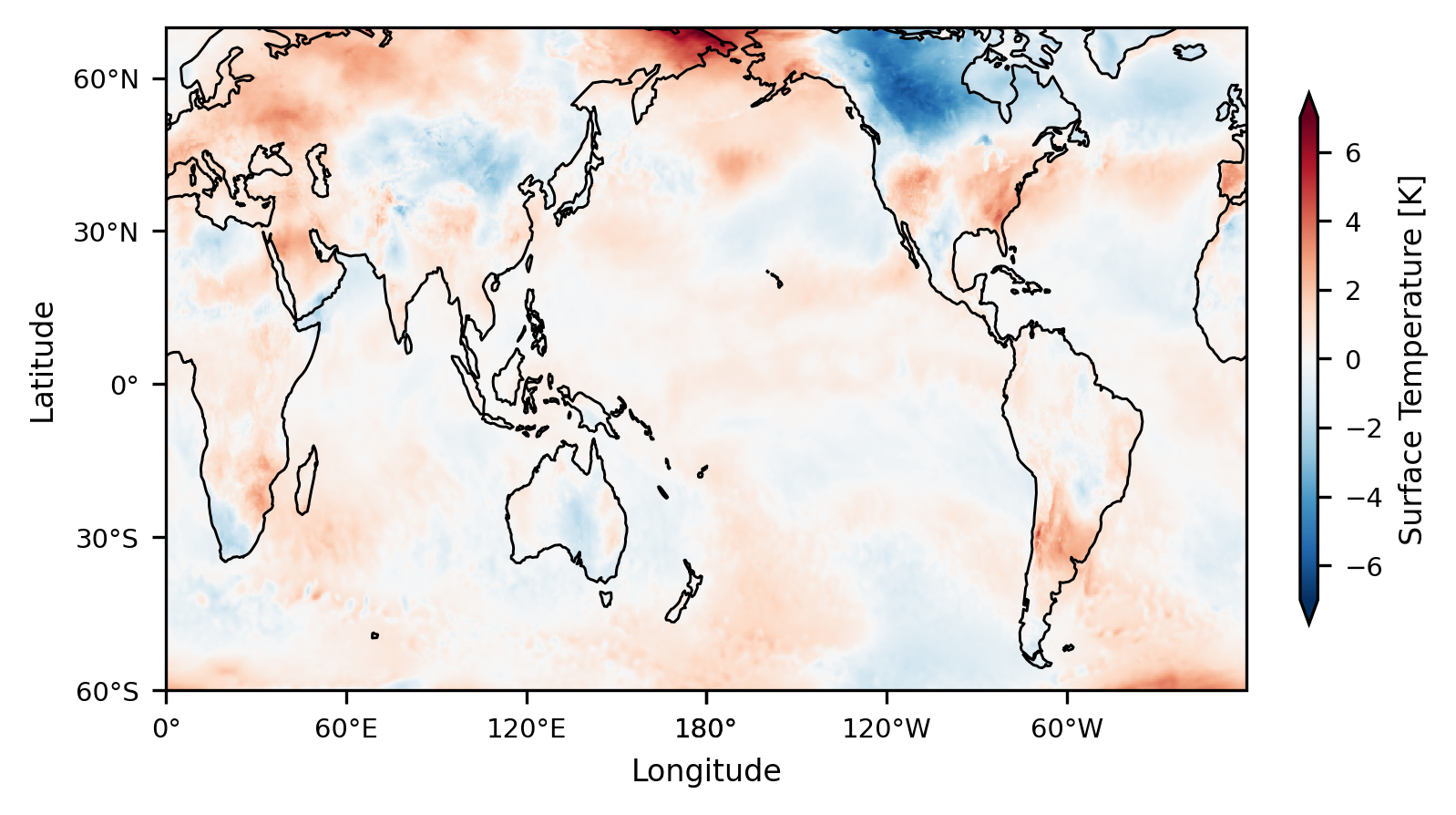}\par 
\vspace{4pt}
\raggedright
下图显示了全球地表温度异常。请分别选择温度异常最高和最低的大致区域。
\begin{itemize}
  \item 白令海
  \item 澳大利亚
  \item 中亚
  \item 北美北部
\end{itemize}

\end{QuestionBox}

\begin{AnswerBox}{Answer}
AD
\end{AnswerBox}

\subsubsection*{Innovator-VL-8B-instruct} 

\begin{ThinkingBox}{Model Output}
AD
\end{ThinkingBox}

\subsubsection*{InternVL3.5-8B} 
\begin{ThinkingBox}{Model Output}
CD
\end{ThinkingBox}

\subsubsection*{MiMo-VL-7B-RL} 
\begin{ThinkingBox}{Model Output}
A
\end{ThinkingBox}

\end{CJK}
\subsection*{Case 11} 

\begin{QuestionBox}{Question}
\centering
\includegraphics[width=0.8\linewidth]{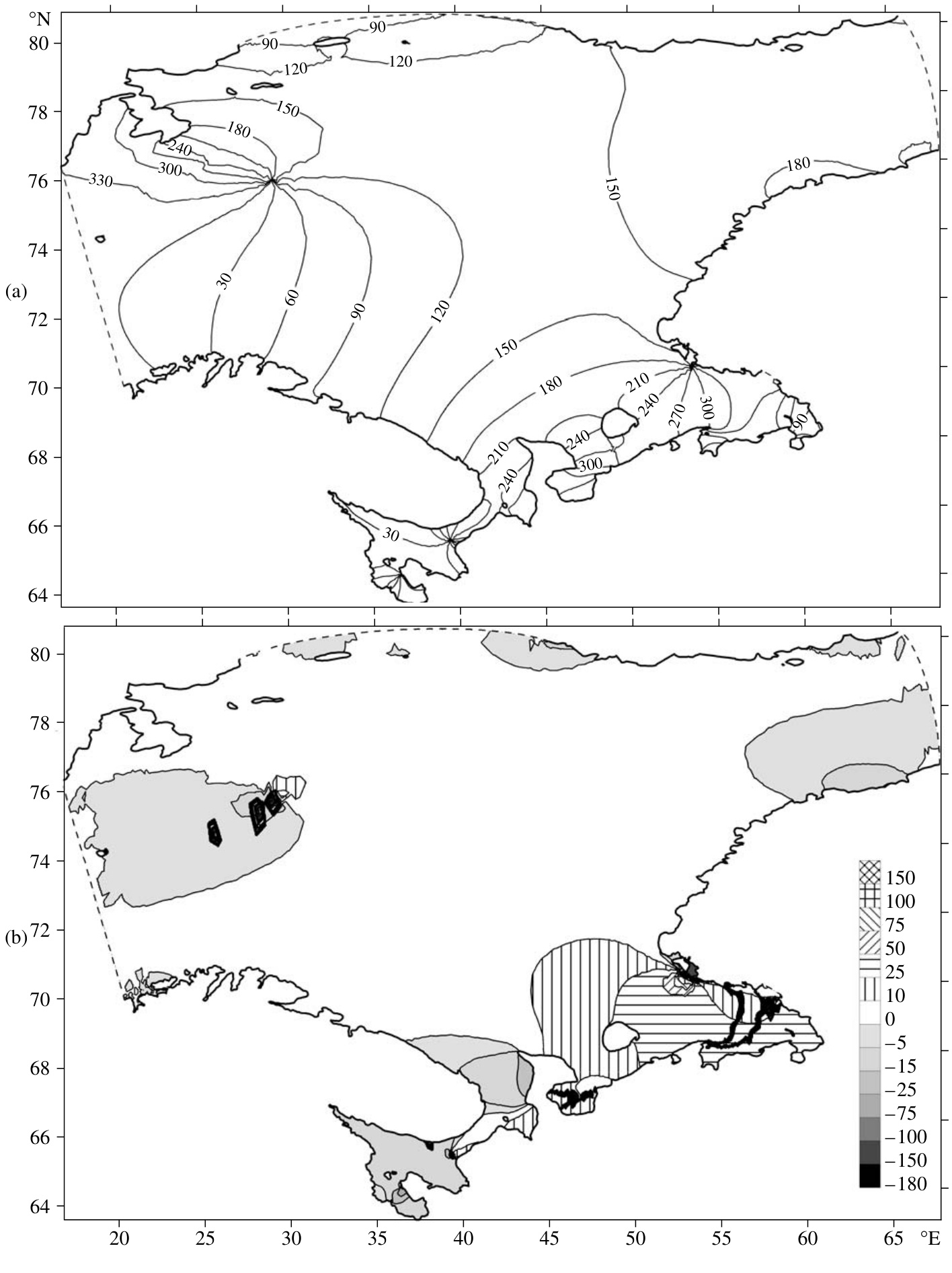}\par 
\vspace{4pt}
\raggedright
You are tasked with answering a multiple-choice question about the given input image.

\medskip
\noindent\textbf{Caption:}
(a) Phases (in degrees) of tidal sea surface level elevations in the ice-free period and (b) their variations (in degrees) under ice or ice-free conditions in the White Sea.

\medskip
\noindent\textbf{Question:}
Which region exhibits degenerate amphidromy characteristics during the ice-free period based on Figure (a)?

\medskip
\noindent\textbf{Options:}
\begin{enumerate}[label=\Alph*.]
  \item Near the throat outlet
  \item Near Cape Letni Orlov
  \item In the northern section of the White Sea
  \item Along the southern coastline
\end{enumerate}

\medskip
Based on the image, select the correct option (e.g., \texttt{A}, \texttt{B}, \texttt{C}) or directly state the correct option content.

\medskip
The output must be written in \textbf{JSON format} using the structure below:
\begin{quote}\ttfamily
\{\newline
\hspace*{1.5em}"answer": "Correct option or short answer",\newline
\hspace*{1.5em}"Explanation": "Reasoning explaining how to derive the correct answer."\newline
\}
\end{quote}

\end{QuestionBox}

\begin{AnswerBox}{Answer}
B. Near Cape Letni Orlov
\end{AnswerBox}

\subsubsection*{Innovator-VL-8B-Instruct} 

\begin{ThinkingBox}{Model Output}
\begin{quote}\ttfamily
\{\newline
\hspace*{1.5em}"answer": "B",\newline
\hspace*{1.5em}"Explanation": "The image shows that near Cape Letni Orlov, the tidal sea surface level elevations exhibit degenerate amphidromy characteristics during the ice-free period."\newline
\}
\end{quote}

\end{ThinkingBox}

\subsubsection*{Qwen3-VL-8B} 
\begin{ThinkingBox}{Model Output}
\begin{quote}\ttfamily
\{\newline
\hspace*{1.5em}"answer": "C. In the northern section of the White Sea",\newline
\hspace*{1.5em}"Explanation": "In Figure (a), the contour lines representing tidal phase (in degrees) show a pattern where the phase values increase radially outward from a central point in the northern section of the White Sea, indicating a degenerate amphidromy. This is characteristic of a single amphidromic point with a radial phase progression, which is typical in the northern part of the White Sea during the ice-free period."\newline
\}
\end{quote}

\end{ThinkingBox}

\subsubsection*{MiMo-VL-7B-RL} 
\begin{ThinkingBox}{Model Output}
\begin{quote}\ttfamily
\{\newline
\hspace*{1.5em}"answer": "C",\newline
\hspace*{1.5em}"Explanation": "In Figure (a), the northern section of the White Sea shows degenerate amphidromy characteristics as indicated by the phase lines converging towards a central point, which is typical of degenerate amphidromy."\newline
\}
\end{quote}

\end{ThinkingBox}

\section{Evaluation Details}
\label{sec:evaluation_details}
\subsection{Benchmark Details.}
\label{Details dataset}

\begin{itemize}
    \item \textbf{General Vision:}
    \begin{itemize}

        \item \textbf{AI2D:}  
        AI2D~\citep{AI2D} contains over 5,000 charts and 15,000 questions and answers, focusing on the syntactic parsing and question-answering tasks for charts. It uses Diagram Parse Graphs (DPG) to support the understanding and reasoning of the structure of charts and their constituent relationships.

        \item \textbf{OCRBench:}  
        OCRBench~\citep{OCRBench} is a comprehensive evaluation benchmark consisting of 29 datasets, aimed at assessing the performance of large multimodal models in text-related visual tasks. It covers a wide range of text recognition tasks, including text recognition, scene text-based question answering, document-based question answering, key information extraction, and handwritten mathematical expression recognition, providing diverse scenarios and tasks to comprehensively evaluate OCR capabilities.

        \item \textbf{ChartQA:}  
        ChartQA~\citep{ChartQA} includes 9.6K human-written questions and 23.1K questions generated from human-written chart summaries, focusing on complex reasoning tasks for charts. Unlike existing template-based question sets, the tasks in this dataset involve multiple logical and arithmetic operations, and the questions often refer to the visual features of the charts. These questions assess the visual and logical reasoning abilities of models and challenge their capacity to answer complex reasoning questions.

        \item \textbf{MMMU:}  
        MMMU~\citep{MMMU} is a large-scale multimodal benchmark with 11.5K questions spanning 6 core disciplines, 30 subjects, 183 subfields, and 30 image types. It is designed to test advanced perception and domain-specific reasoning abilities.

        \item \textbf{MMMU-Pro:}  
        MMMU-Pro~\citep{MMMU} is an enhanced version of the MMMU benchmark, which rigorously evaluates the understanding and reasoning capabilities of multimodal models through a three-step evaluation process: (1) filtering out questions answerable by text-only models, (2) augmenting candidate options, and (3) introducing a vision-only input setting where questions are embedded within images. This setting tests models on their ability to process both visual and textual information simultaneously, and results show that models perform significantly worse on MMMU-Pro compared to MMMU.
        
        \item \textbf{MMStar:}  
        MMStar~\citep{MMstar} is a vision-indispensable benchmark with 1,500 human-curated items balanced over six core capabilities and 18 axes. It explicitly filters out samples that can be solved without images and proposes metrics to quantify the true multimodal gains.
        
        \item \textbf{VStar-Bench:}  
        VStar-Bench~\citep{VStar-Bench} is a benchmark specifically designed to evaluate large multimodal language models (MLLMs) in their ability to process high-resolution images and focus on visual details. The benchmark introduces a V-visual search mechanism to test MLLMs' performance in visual querying, collaborative reasoning, and contextual understanding, emphasizing the importance of incorporating visual search capabilities into multimodal systems.

        \item \textbf{MMBench:}  
        MMBench~\citep{MMBench} is a systematically designed bilingual multimodal evaluation benchmark that covers a wide range of fine-grained vision–language capabilities. Through rigorous data quality control and a cyclic evaluation strategy, it provides robust and objective model assessment. The benchmark supports both Chinese and English multiple-choice evaluations and introduces large language model-assisted grading mechanisms, enabling fair and scalable comparison across models with different instruction-following capabilities.
        
        \item \textbf{MME-RealWorld:}  
        MME-RealWorld~\citep{MME-realworld} is a large-scale, human-annotated multimodal benchmark focusing on high-resolution images and complex real-world scenarios. It systematically characterizes the core bottlenecks of current MLLMs in perception and reasoning. By covering diverse scenes and subtasks with challenging, manually constructed questions, the benchmark reveals significant shortcomings even in state-of-the-art models under realistic application settings.
        
        \item \textbf{DocVQA:}  
        DocVQA~\citep{DocVQA} is a document image-based visual question answering dataset containing over 12,000 document images and 50,000 questions. It is designed to systematically evaluate models’ capabilities in document understanding, layout reasoning, and structural perception.
        
        \item \textbf{InfoVQA:}  
        InfoVQA~\citep{InfoVQA} targets infographic understanding and visual question answering, covering diverse information graphics annotated with natural language questions and answers. The dataset requires joint reasoning over document layout, textual content, graphical elements, and data visualizations, emphasizing foundational reasoning and arithmetic skills in complex infographic comprehension.
        
        \item \textbf{SEED-Bench:}  
        SEED-Bench~\citep{SEED-Bench} is a benchmark for evaluating the understanding capabilities of generative multimodal large language models. It consists of 19,000 human-annotated multiple-choice questions spanning 12 key understanding dimensions across both image and video modalities.

        \item \textbf{SEED-Bench-2-Plus:}
        SEED-Bench-2-Plus~\citep{seed-banch2} is a benchmark for evaluating text-rich visual comprehension of multimodal large language models, consisting of 2.3K human-annotated multiple-choice questions. The dataset covers three categories—Charts, Maps, and Webs—to assess model performance in real-world scenarios involving dense embedded text in images.
        
        \item \textbf{RealWorldQA:}  
        RealWorldQA~\citep{RealWorldQA} contains 765 images annotated with question–answer pairs, covering anonymized vehicle-captured data and a wide range of real-world scenes. It aims to evaluate multimodal models under realistic and unconstrained visual environments.
        
    \end{itemize}
    
    \item \textbf{Math \& Reasoning:}
    \begin{itemize}
        
        \item \textbf{MathVista:}  
        MathVista~\citep{MathVista} targets mathematical reasoning in visual contexts, aggregating 6,141 examples from 28 existing datasets and 3 newly created ones, testing models’ abilities in fine-grained visual understanding and compositional reasoning.

        \item \textbf{MathVision:}  
        MathVision~\citep{MathVision} curates 3,040 competition-style math problems with visual contexts across 16 disciplines and five difficulty levels, highlighting gaps between current models and human performance in multimodal mathematical reasoning.

        \item \textbf{MathVerse:}  
        MathVerse~\citep{MathVerse} is a comprehensive visual mathematics benchmark for multimodal large language models, consisting of 2,612 image-based math problems. It introduces multiple modality variants to fairly assess whether models truly leverage visual information rather than relying solely on textual reasoning. The benchmark further adopts fine-grained chain-of-thought-based evaluation strategies to analyze intermediate reasoning processes in depth.
        
        \item \textbf{WeMath:}  
        WeMath~\citep{qiao2025we} is the first benchmark focusing on knowledge acquisition and generalization mechanisms in visual mathematical reasoning. It contains 6.5K problems covering 67 hierarchical mathematical knowledge concepts. By decomposing questions into subproblems and introducing four-dimensional evaluation metrics, WE-MATH systematically reveals the fundamental capabilities and limitations of multimodal models in understanding and generalizing mathematical knowledge.

    \end{itemize}
    
    \item \textbf{Science:}
    \begin{itemize}
        \item \textbf{ScienceQA:}  
        ScienceQA~\citep{ScienceQA} is a large-scale multimodal science question-answering benchmark containing approximately 21,000 multiple-choice questions across diverse scientific disciplines. Each question is accompanied by detailed explanations and reasoning chains, explicitly designed to characterize models’ chain-of-thought reasoning, interpretability, and generalization in multi-hop multimodal scientific reasoning.
        
        \item \textbf{RxnBench:}  
        RxnBench~\citep{li2025rxnbench} is a multimodal benchmark for chemical reaction understanding, constructed from real scientific paper PDFs. It evaluates models' joint understanding of reaction diagrams and textual descriptions through both single-image QA and full-document QA tasks, systematically exposing limitations in chemical structure recognition and mechanistic reasoning.
        
        \item \textbf{MolParse:}
        Mol QA \& Parser is an evaluation benchmark for molecular understanding, containing 2,000 question–answer pairs based on single-molecule images. It is used to evaluate molecular question answering and structure parsing tasks, including functional group recognition, property-related queries, and SMILES or extended SMILES generation from molecular images.
        
        \item \textbf{OpenRxn:}
        Reaction is an evaluation benchmark for chemical reaction understanding, containing 2,000 reaction-based question–answer pairs derived from organic reaction images and SMILES representations. It is used to evaluate chemical knowledge and reaction-level reasoning in large language models.      
        
        \item \textbf{EMVista:}  
        A benchmark of 1,850 EM microstructure samples evaluating instance-level perception, microstructural attribute understanding (morphology, density, spatial distribution, layering, scale variation), and robustness under dense, overlapping, multi-scale scenes.

        \item \textbf{SmolInstruct:}  
        SMolInstruct~\citep{smolinstruct} is a large-scale benchmark for evaluating chemistry-oriented language models, covering 14 small-molecule tasks and approximately 3.3M question–answer pairs with standardized evaluation splits.
        
        \item \textbf{SuperChem:}  
        SUPERChem~\citep{zhao2025superchem} is a 500-problem, expert-curated chemistry reasoning benchmark (multimodal + text-only) with expert solution paths for process-level evaluation via Reasoning Path Fidelity (RPF).

        \item \textbf{ProteinLMBench:}  
        ProteinLMBench~\citep{shen2024fine} consists of 944 manually verified multiple-choice questions covering protein sequence, structure, and function. It is specifically designed for evaluating protein understanding in LLMs.

        \item \textbf{SFE:}  
        Scientists' First Exam (SFE)~\citep{zhou2025scientists} probes scientific cognition of MLLMs across perception, attribute understanding, and comparative reasoning, comprising 830 expert-verified VQA items across five disciplines.

        \item \textbf{MicroVQA:}  
        MicroVQA~\citep{MicroVQA} is a microscopy-centric multimodal benchmark with 1,042 multiple-choice questions spanning diverse imaging modalities and biological topics, targeting scientific analysis and reasoning.

        \item \textbf{MSEarth-MCQ:}  
        MSEarth-MCQ~\citep{MSEarth} is the multiple-choice subset of the MSEarth Earth-science benchmark, providing approximately 2.78K expert-derived figure-grounded questions across the atmosphere, cryosphere, hydrosphere, and biosphere.

        \item \textbf{XLRS-Bench:}  
        XLRS-Bench~\citep{XLRS-Bench-lite} evaluates MLLMs on ultra-high-resolution remote-sensing imagery (average resolution ~8.5k × 8.5k pixels), with human-verified annotations across diverse tasks reflecting real-world RS scenarios.
    \end{itemize}
\end{itemize}

\subsection{Evaluation Prompts}


\begin{UserPromptBox}{AI2D}
\texttt{<image>} \\
\{question\} \\
A. \{choice\_1\} \\
B. \{choice\_2\} \\
C. \{choice\_3\} \\
D. \{choice\_4\} \\
Answer with the option's letter from the given choices directly.
\end{UserPromptBox}

\begin{UserPromptBox}{AI2D (no mask)}
\texttt{<image>} \\
\{question\} \\
A. \{choice\_1\} \\
B. \{choice\_2\} \\
C. \{choice\_3\} \\
D. \{choice\_4\} \\
Answer with the option's letter from the given choices directly.
\end{UserPromptBox}

\begin{UserPromptBox}{OCRBench}
\texttt{<image>} \\
\{question\}
\end{UserPromptBox}

\begin{UserPromptBox}{ChartQA}
\texttt{<image>} \\
\{question\} \\
Answer the question with a single word.
\end{UserPromptBox}

\begin{UserPromptBox}{MMMU (Val)}
\textbf{multiple\_choice\_prompt:} \\
\texttt{<image>} \\
\{question\} \\
A. \{option\_1\} \\
B. \{option\_2\} \\
C. \{option\_3\} \\
D. \{option\_4\} \\
Answer with the option's letter from the given choices directly.

\vspace{10pt} 

\textbf{open\_ended\_prompt:} \\
\texttt{<image>} \\
\{question\} \\
Answer the question using a single word or phrase.
\end{UserPromptBox}

\begin{UserPromptBox}{MMMU-Pro (Standard)}
\texttt{<image>} \\
\{question\} \\
A. \{choice\_1\} \\
B. \{choice\_2\} \\
... \\
J. \{choice\_10\} \\
Answer with the option letter from the given choices directly.
\end{UserPromptBox}

\begin{UserPromptBox}{MMStar}
\texttt{<image>} \\
\{question\} \\
Answer with the option's letter from the given choices directly.
\end{UserPromptBox}

\begin{UserPromptBox}{VStar-Bench}
\texttt{<image>} \\
\{question\} \\
A. \{option\_1\} \\
B. \{option\_2\} \\
C. \{option\_3\} \\
D. \{option\_4\} \\
Answer with the option's letter from the given choices directly.
\end{UserPromptBox}

\begin{UserPromptBox}{MMBench-EN (Dev/Test)}
\texttt{<image>} \\
\{hint\} \{question\} \{options\} \\
Answer with the option's letter from the given choices directly.
\end{UserPromptBox}

\begin{UserPromptBox}{MME-RealWorld (EN)}
\texttt{<image>} \\
\{question\} The choices are listed below: \\
\{options\} \\
Select the best answer to the above multiple-choice question based on the image. Respond with only the letter (A, B, C, D, or E) of the correct option. \\
The best answer is:
\end{UserPromptBox}

\begin{CJK}{UTF8}{gbsn} 
\begin{UserPromptBox}{MME-RealWorld (CN)}
\texttt{<image>} \\
\{question\} 选项如下所示: \\
\{options\} \\
根据图像选择上述多项选择题的最佳答案。只需回答正确选项的字母（A, B, C, D 或 E）。 \\
最佳答案为：
\end{UserPromptBox}
\end{CJK} 

\begin{UserPromptBox}{DocVQA(Val)}
\texttt{<image>} \\
\{question\} \\
Answer the question using a single word or phrase.
\end{UserPromptBox}

\begin{UserPromptBox}{InfoVQA(Val)}
\texttt{<image>} \\
\{question\} \\
Answer the question using a single word or phrase.
\end{UserPromptBox}

\begin{UserPromptBox}{SEED-Bench / SEED-Bench-2-Plus}
\texttt{<image>} \\
\{question\} \\
A. \{choice\_A\} ... D. \{choice\_D\} \\
Answer with the option's letter from the given choices directly.
\end{UserPromptBox}

\begin{UserPromptBox}{RealWorldQA}
\texttt{<image>} \\
\{question\}
\end{UserPromptBox}


\begin{UserPromptBox}{MathVision / MathVision-mini}
Think and solve the following question step by step. Please put your thinking and analysis procedure within \texttt{<think></think>}. Put ONLY your final answer within \texttt{<answer></answer>}. \\
\texttt{<image>} \\
\{question\} \\
Choices: \{choices\}
\end{UserPromptBox}

\begin{UserPromptBox}{MathVerse}
Think and solve the following question step by step. Please put your thinking and analysis procedure within \texttt{<think></think>}. Put ONLY your final answer within \texttt{<answer></answer>}. \\
\texttt{<image>} \\
\{question\}
\end{UserPromptBox}

\begin{UserPromptBox}{MathVista}
Think and solve the following question step by step. Please put your thinking and analysis procedure within \texttt{<think></think>}. Put ONLY your final answer within \texttt{<answer></answer>}. \\
Question: \{question\} (Unit: \{unit\}) \\
Choices: \{choices\} \\
Hint: \{hint\}
\end{UserPromptBox}

\begin{UserPromptBox}{WeMath}
Think and solve the following question step by step. Please put your thinking and analysis procedure within \texttt{<think></think>}. Put ONLY your final answer within \texttt{<answer></answer>}. \\
\texttt{<image>} \\
\{question\} \\
\{option\}
\end{UserPromptBox}


\begin{UserPromptBox}{ScienceQA}
\texttt{<image>} \\
Context: \{hint\} \\
\{question\} \\
\{choices\} \\
Answer with the option's letter from the given choices directly.
\end{UserPromptBox}


\begin{UserPromptBox}{RxnBench (EN)}
\texttt{<image>} \\
Question: \{question\} \\
Choices: \\
A. \{choice\_A\} \\
B. \{choice\_B\} \\
C. \{choice\_C\} \\
D. \{choice\_D\} \\
Based on the image and the question, choose the most appropriate answer. \\
\textbf{Only output a single letter (A, B, C, or D)}. Do NOT output any other text or explanation.
\end{UserPromptBox}

\begin{CJK}{UTF8}{gbsn} 
\begin{UserPromptBox}{RxnBench (ZH)}
\texttt{<image>} \\
问题: \{question\} \\
选项: \\
A. \{choice\_A\} \\
B. \{choice\_B\} \\
C. \{choice\_C\} \\
D. \{choice\_D\} \\
请根据图像和问题，从以上四个选项中选择最合适的答案。 \\
\textbf{只输出单个字母（A, B, C 或 D），不要输出选项内容，也不要输出任何解释。}
\end{UserPromptBox}
\end{CJK} 


\begin{UserPromptBox}{MolParse}
\texttt{<image>} \\
\{question\}
\end{UserPromptBox}

\begin{UserPromptBox}{OpenRxn}
\texttt{<image>} \\
\{question\}
\end{UserPromptBox}


\begin{UserPromptBox}{EMVista}
\texttt{<image>} \\
\{problem\} \\
Answer with the option's letter from the given choices directly.
\end{UserPromptBox}

\begin{UserPromptBox}{SuperChem (EN)}
\texttt{<image>} \\
Question:\\
\{question\} \\
\{option\} \\
Select the best answer to the above multiple-choice question based on the image. Respond with only the letter of the correct option.
\end{UserPromptBox}

\begin{CJK}{UTF8}{gbsn} 
\begin{UserPromptBox}{SuperChem (CN)}
\texttt{<image>} \\
问题: \\
\{question\} \\
\{option\} \\
根据图像选择上述单选题的最佳答案。请仅输出正确选项的字母。 \\
\end{UserPromptBox}
\end{CJK} 


\begin{UserPromptBox}{ProteinLMBench}
Answer the multiple-choice question based solely on the provided context. If you are still unsure about the answer, output option 7. \\
Select only ONE correct option by its number. Start your response with 'The correct option is' followed by the option number ONLY. \\
Question: \{question\} \\
Options: \{options\} \\
The correct option is:
\end{UserPromptBox}

\begin{UserPromptBox}{SFE-EN / SFE-ZH}
\textbf{multiple\_choice\_prompt:} \\
\texttt{<image>} \\
You are an expert in \{field\} and need to solve the following question. \\
The question is a multiple-choice question. Answer with the option letter from the given choices. \\
\{question\} \\
\{options\}

\vspace{8pt} 

\textbf{exact\_match\_prompt:} \\
\texttt{<image>} \\
You are an expert in \{field\} and need to solve the following question. \\
The question is an exact match question. Answer the question using a single word or phrase. \\
\{question\}

\vspace{8pt} 

\textbf{open\_ended\_prompt:} \\
\texttt{<image>} \\
You are an expert in \{field\} and need to solve the following question. \\
The question is an open-ended question. Answer the question using a phrase. \\
\{question\}
\end{UserPromptBox}

\begin{UserPromptBox}{MicroVQA}
\texttt{<image>} \\
The following is a multiple choice question (with answers). \\ 
Think step by step and then output the answer in the format of "The answer is (X)" at the end, where X is the correct letter choice. \\
\{question\} \\
Options: \{options\}
\end{UserPromptBox}

\begin{UserPromptBox}{MSEarth-MCQ}
\texttt{<image>} \\
You are tasked with answering a multiple-choice question about the given input image. \\
\{question\} \\
Based on the image, select the correct option (e.g., 'A', 'B', 'C') or directly state the correct option content. \\
The output must be written in JSON format using the structure below: \\
```json \\
\{\{ \\ 
    "answer": "Correct option or short answer", \\
    "Explanation": "Reasoning explaining how to derive the correct answer." \\
\}\} \\
```
\end{UserPromptBox}
















\begin{UserPromptBox}{SmolInstruct}

\textbf{HIV} \\
Analyze the SMILES string and predict its HIV activity. 
Answer strictly with ``Yes'' or ``No''. \\
Input: \{input\} \\
Output:

\vspace{6pt}

\textbf{BBBP} \\
Does this molecule penetrate the Blood-Brain Barrier (BBBP)?
Answer strictly with ``Yes'' or ``No''. \\
Input: \{input\} \\
Output:

\vspace{6pt}

\textbf{ClinTox} \\
Predict the clinical toxicity of this molecule.
Answer strictly with ``Yes'' or ``No''. \\
Input: \{input\} \\
Output:

\vspace{6pt}

\textbf{Side Effect} \\
Does this drug cause the specific side effect?
Answer strictly with ``Yes'' or ``No''. \\
Input: \{input\} \\
Output:

\vspace{8pt}

\textbf{Aqueous Solubility} \\
As a specialized chemist, calculate the logSol (aqueous solubility) for the molecule below.
Consider the molecular weight, number of rotatable bonds, and aromatic proportion.
Provide the final logSol value as a floating point number. Output: [value] \\
Input: \{input\} \\
Output:

\vspace{6pt}

\textbf{Lipophilicity} \\
Analyze the lipophilicity (logP) of this molecular structure.
Focus on the hydrophobic and hydrophilic balance.
Provide the numerical logP value only. Output: [value] \\
Input: \{input\} \\
Output:

\vspace{8pt}

\textbf{Forward Synthesis} \\
Predict the major product(s) for these reactants.
Output the product(s) in SMILES format. \\
Input: \{input\} \\
Output:

\end{UserPromptBox}

\begin{UserPromptBox}{SmolInstruct}

\textbf{Retrosynthesis} \\
Suggest the starting materials (reactants) for the given product.
If multiple reactants are needed, separate them with a period (.).
Output only the SMILES strings. \\
Input: \{input\} \\
Output:

\vspace{8pt}

\textbf{SMILES to IUPAC} \\
Convert this SMILES structure to its IUPAC name.
Output only the IUPAC name, without any explanation or extra text. \\
Input: \{input\} \\
Output:

\vspace{6pt}

\textbf{IUPAC to SMILES} \\
Convert this IUPAC name to its SMILES structure.
Output only the SMILES string, without any explanation or extra text. \\
Input: \{input\} \\
Output:

\vspace{6pt}

\textbf{SMILES to Molecular Formula} \\
Determine the molecular formula of this SMILES string.
Output only the molecular formula, without any explanation or extra text. \\
Input: \{input\} \\
Output:

\vspace{6pt}

\textbf{IUPAC to Molecular Formula} \\
Determine the molecular formula of this IUPAC name.
Output only the molecular formula, without any explanation or extra text. \\
Input: \{input\} \\
Output:

\vspace{8pt}

\textbf{Captioning} \\
Describe the following molecule's structure and chemical classification in detail. \\
Input: \{input\} \\
Output:

\vspace{6pt}

\textbf{Generation} \\
Generate the SMILES string for a molecule that fits this description. \\
Input: \{input\} \\ 
Output:

\end{UserPromptBox}

\begin{UserPromptBox}{XLRS-Bench-lite}
\textbf{w/o Land Use Classification:} \\
\texttt{<image>} \\
\{question\} The choices are listed below: \\
\{option\_A\} \\
\{option\_B\} \\
\{option\_C\} \\
\{option\_D\} \\
Select the best answer for the multiple-choice question based on the image. Only respond with the letter corresponding to the correct answer (A, B, C, D). \\
The answer is:

\vspace{10pt} 

\textbf{w/ Land Use Classification:} \\
\texttt{<image>} \\
\{question\} The choices are listed below: \\
\{option\_A\} \\
\{option\_B\} \\
\{option\_C\} \\
\{option\_D\} \\
Select the best answer(s) for the multiple-choice question based on the image. There may be more than one correct option. Only respond with the letter(s) corresponding to the correct answer(s) (A, B, C, D), with multiple choices separated by spaces. The answer(s) is(are):
\end{UserPromptBox}

\end{document}